\documentclass[sigconf]{acmart}

\usepackage{overpic}
\usepackage{tcolorbox}
\usepackage{colortbl} 
\usepackage{enumitem}

\AtBeginDocument{%
  \providecommand\BibTeX{{%
    \normalfont B\kern-0.5em{\scshape i\kern-0.25em b}\kern-0.8em\TeX}}}

\copyrightyear{2024}
\acmYear{2024}
\setcopyright{rightsretained}
\acmConference[MM '24]{Proceedings of the 32nd ACM International Conference
on Multimedia}{October 28-November 1, 2024}{Melbourne, VIC, Australia}
\acmBooktitle{Proceedings of the 32nd ACM International Conference on
Multimedia (MM '24), October 28-November 1, 2024, Melbourne, VIC, Australia}
\acmDOI{10.1145/3664647.3680665}
\acmISBN{979-8-4007-0686-8/24/10}

\begin{document}

\title{AutoM\textsuperscript{3}L: 
An Automated Multimodal Machine Learning Framework with Large Language Models
}




\author{Daqin Luo}
\affiliation{%
 \institution{Dataa Robotics}
 \city{Shenzhen}
 \country{China}}
\email{daqinluo@outlook.com}

\author{Chengjian Feng}
\affiliation{%
  \institution{Meituan Inc.}
  \city{Shenzhen}
  \country{China}}
\email{fcjian@outlook.com}

\author{Yuxuan Nong}
\affiliation{%
  \institution{Dataa Robotics}
  \city{Shenzhen}
  \country{China}
  }
\email{yuxuan.nong@dataarobotics.com}

\author{Yiqing Shen}
\authornote{Corresponding author.}
\affiliation{%
  \institution{Johns Hopkins University}
  \city{Baltimore}
  \country{USA}}
\email{yshen92@jhu.edu}





\renewcommand{\shortauthors}{Daqin Luo and Chengjian Feng, et al.}

\begin{abstract}
  Automated Machine Learning (AutoML) offers a promising approach to streamline the training of machine learning models.
  However, existing AutoML frameworks are often limited to unimodal scenarios and require extensive manual configuration.
  Recent advancements in Large Language Models (LLMs) have showcased their exceptional abilities in reasoning, interaction, and code generation, presenting an opportunity to develop a more automated and user-friendly framework.
  To this end, we introduce \texttt{AutoM\textsuperscript{3}L}, an innovative Automated Multimodal Machine Learning framework that leverages LLMs as controllers to automatically construct multimodal training pipelines. 
  \texttt{AutoM\textsuperscript{3}L} comprehends data modalities and selects appropriate models based on user requirements, providing automation and interactivity. 
  By eliminating the need for manual feature engineering and hyperparameter optimization, our framework simplifies user engagement and enables customization through directives, addressing the limitations of previous rule-based AutoML approaches.
  We evaluate the performance of \texttt{AutoM\textsuperscript{3}L} on six diverse multimodal datasets spanning classification, regression, and retrieval tasks, as well as a comprehensive set of unimodal datasets.
  The results demonstrate that \texttt{AutoM\textsuperscript{3}L} achieves competitive or superior performance compared to traditional rule-based AutoML methods.
  Furthermore, a user study highlights the user-friendliness and usability of our framework, compared to the rule-based AutoML methods.
  Code is available at:
    \href{https://github.com/tim120526/AutoM3L}{\textcolor{magenta}{https://github.com/tim120526/AutoM3L}}. 
\end{abstract}




\maketitle

\begin{figure*}[ht!]
    \centering
    \includegraphics[width=0.86\linewidth]{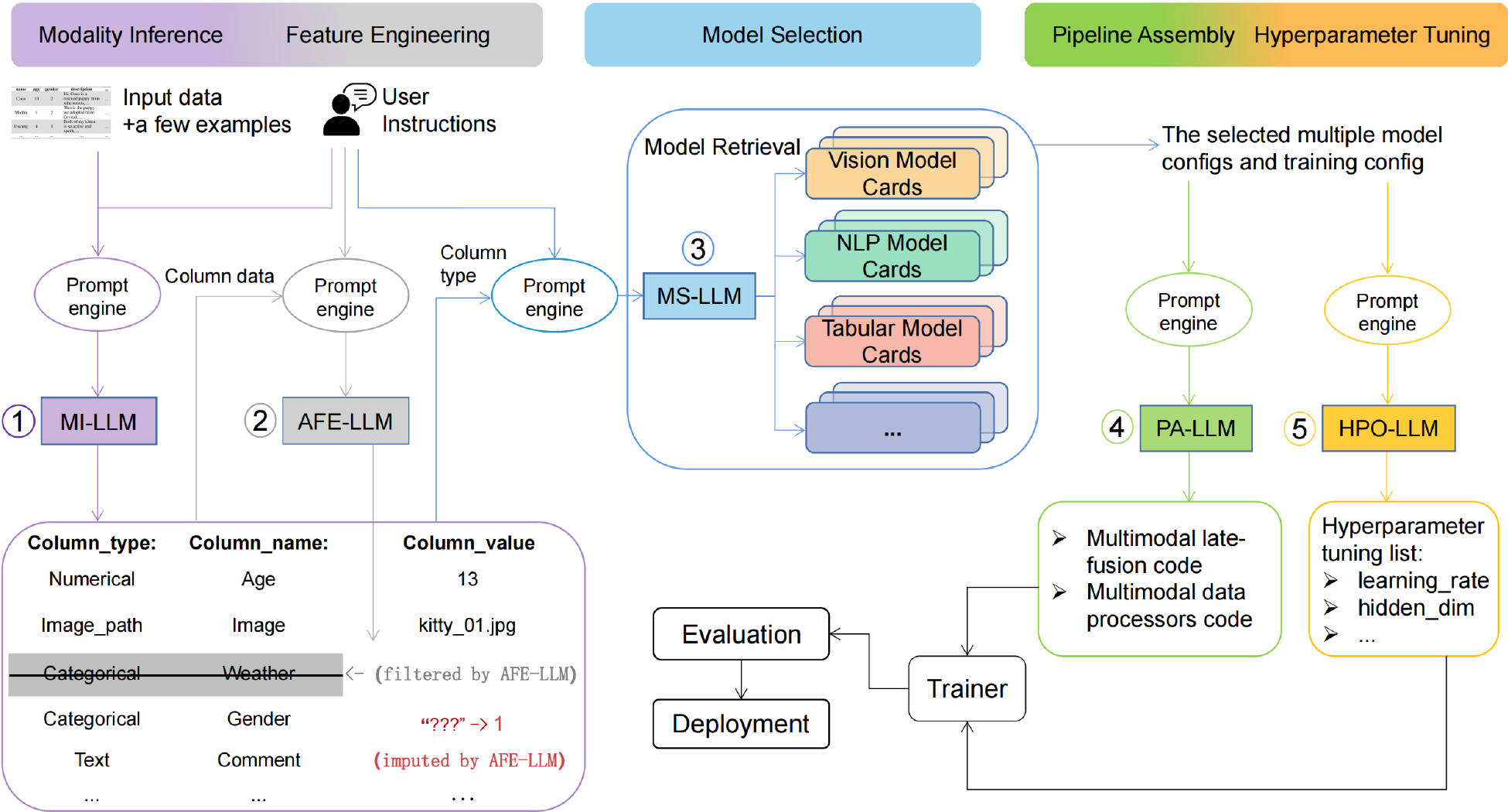}
    \caption{
    The overall framework of \texttt{AutoM\textsuperscript{3}L}. It consists of five stages: 
    \textcircled{\raisebox{-0.1ex}{\scalebox{0.9}{1}}} Infer the modality of each attribute in structured table data. 
    \textcircled{\raisebox{-0.1ex}{\scalebox{0.9}{2}}} Automate feature engineering for feature filtering and data imputation. 
    \textcircled{\raisebox{-0.1ex}{\scalebox{0.9}{3}}} Select optimal models for each modality. 
    \textcircled{\raisebox{-0.1ex}{\scalebox{0.9}{4}}} Generates executable scripts for model fusion and data processing to assemble the training pipeline. 
    \textcircled{\raisebox{-0.1ex}{\scalebox{0.9}{5}}} Search optimal hyperparameters.
    The detailed system prompts for LLMs in each stage can be found in Appendix~\ref{appendix-prompts}.}
    \label{fig1}
\end{figure*}

\section{Introduction}
Multimodal data is crucial in various machine learning (ML) tasks as it provides the ability to capture more comprehensive feature representations.
Real-world data often combines heterogeneous sources, such as integrating table product information with associated images and textual descriptions.
Similarly, in the financial sector, user photos, text, transactions, and other data types are frequently consolidated in tabular formats for analysis and management.
However, the inherent diversity of these modalities introduces complexities, particularly in selecting optimal machine learning or deep learning model architectures and seamlessly synchronizing features across modalities
Consequently, there is often a heavy reliance on manual involvement in the ML pipeline.

Automated Machine Learning (AutoML) has emerged as a promising approach to reduce the need for manual intervention in the ML pipeline \cite{hutter2019automated,gijsbers2019open,vakhrushev2021lightautoml,weerts2020importance,wang2021flaml,elshawi2019automated}. 
However, a significant gap exists for multimodal data, as the majority of AutoML solutions primarily focus on unimodal data. 
AutoGluon\footnote{\url{https://github.com/autogluon/autogluon}} made an initial attempt at multimodal AutoML but suffers from several limitations.
Firstly, it lacks comprehensive automation of feature engineering, which is crucial for effectively handling multimodal data.
Secondly, it presents a steep learning curve for users to become familiar with its configurations and settings, contradicting the user-friendly automation principles that AutoML aims to embody.
Moreover, AutoGluon's adaptability is constrained by pre-set settings such as the search space, model selection, and hyperparameters, necessitating significant manual intervention.
Lastly, extending AutoGluon's capabilities by integrating new techniques or models often requires complex manual code modifications, hindering its agility and potential for growth.

The scientific community has been captivated by the rapid rise of large language models (LLMs), particularly due to their transformative potential in task automation \cite{brown2020language,chowdhery2022palm,touvron2023llama,wei2022emergent}.
LLMs have evolved beyond their initial purpose as text generators and have now become highly autonomous entities capable of self-initiated planning and execution \cite{shen2023hugginggpt,wang2023voyager,wu2023visual,hong2023metagpt,yao2022react}.
This evolution presents a compelling opportunity to enhance the performance and adaptability of multimodal AutoML systems.
Leveraging this potential, we introduce \texttt{AutoM\textsuperscript{3}L}, an innovative LLM framework for Automated Multimodal Machine Learning.
Unlike platforms such as AutoGluon, which are constrained by predefined pipelines, \texttt{AutoM\textsuperscript{3}L} distinguishes itself through its dynamic user interactivity.
Specifically, it seamlessly integrates ML pipelines tailored to user instructions, enabling unparalleled scalability and adaptability throughout the entire process, from data pre-processing to model selection and optimization.

The major contributions are four-fold, summarized as follows.
(1) We introduce AutoM\textsuperscript{3}L, a novel framework that automates the development of machine learning pipelines for multimodal data. 
AutoM\textsuperscript{3}L enables users to derive accurate models for each modality from a diverse pool of models and generates an executable script for cross-modality feature fusion, all with minimal natural language instructions. This approach simplifies the process of building multimodal ML pipelines and makes it more accessible to a wider range of users.
(2) We advance the automation of feature engineering by leveraging a LLM to intelligently filter out attributes that could hinder model performance while simultaneously imputing missing data.
This automated feature engineering process reduces the need for manual intervention and improves the overall quality of the input data.
(3) We automate hyperparameter optimization by combining the LLM's self-generated suggestions with external API calls.
This approach eliminates the need for labor-intensive manual explorations and enables more efficient and effective hyperparameter tuning.
(4) We conduct comprehensive evaluations, comparing AutoM\textsuperscript{3}L with conventional rule-based AutoML on a diverse set of multimodal and unimodal datasets.
Additionally, a user study further highlighted the distinct advantages of our framework in terms of user-friendliness and a significantly reduced learning curve.

\section{Methods}

\label{headings}

In this paper, we propose an \textbf{Auto}mated \textbf{M}ulti-\textbf{M}odal \textbf{M}achine \textbf{L}earning (\texttt{AutoM\textsuperscript{3}L}) framework that utilizes Large Language Models (LLMs) to automate the machine learning pipeline for multimodal scenarios.
This section begins by introducing the organization of the multimodal dataset in Sec.\ref{Organization}. In Sec.\ref{Modality} to \ref{Automated}, we elaborate on the five functional components enhanced by LLMs in \texttt{AutoM\textsuperscript{3}L}: (1) modality inference, (2) automated feature engineering, (3) model selection, (4) pipeline assembly, and (5) hyperparameter optimization, as illustrated in Fig.~\ref{fig1}.

\subsection{Organization of Multimodal Dataset}
\label{Organization}

Most existing studies utilize the JavaScript Object Notation (JSON) to represent multimodal data. However, JSON cannot capture the interplay between different modalities, making it unsuitable for analysis by language models. To address this limitation, we follow \cite{gu-budhkar-2021-package,shi2021benchmarking,hu2024pytorch,burton2023shap} and employ the structured tables to represent multimodal data. Structured tables offer a clear representation that captures the interaction between different modalities and effectively aggregates information from various formats into a unified structure. Additionally, these tables encompass a diverse range of data modalities, including images, text, tabular data, and more.

\begin{figure}[t]
\centering   
\includegraphics[width=\linewidth]{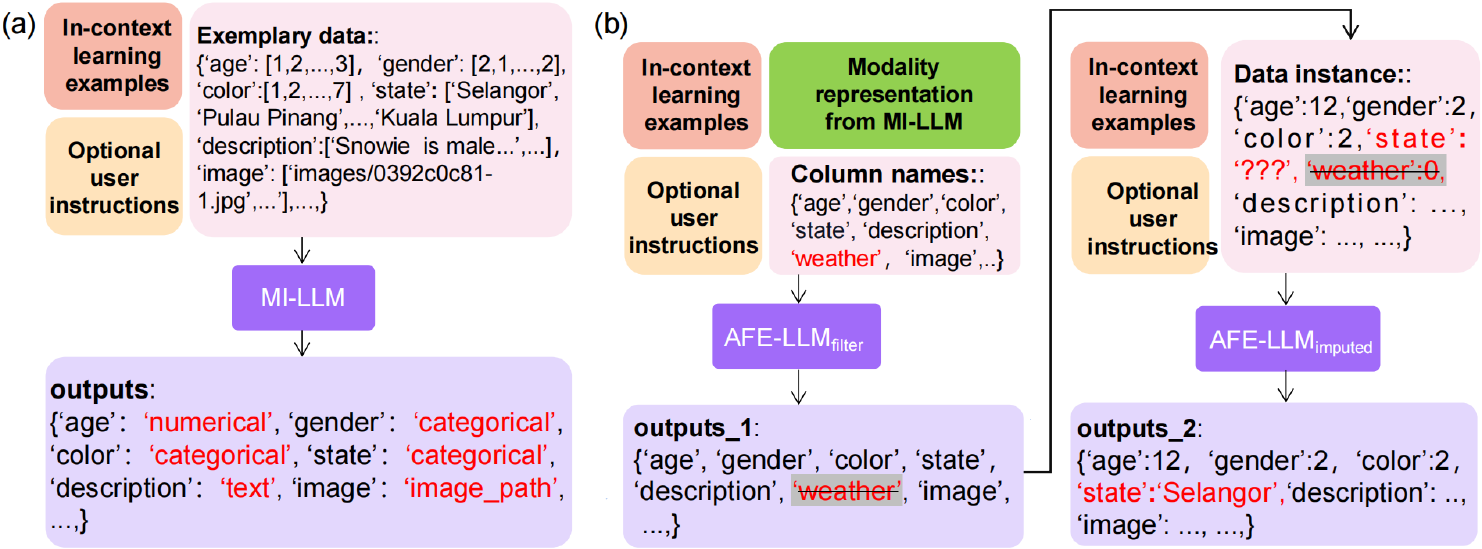}
\caption{
(a) Modality Inference with MI-LLM. It displays MI-LLM's capability to determine the modality of each column in a dataset. Attributes are annotated in red to indicate the inferred modality.
(b) Data Refinement with AFE-LLM. It highlights AFE-LLM's dual role in feature filtering and data imputation. The left part displays attributes marked in red that are filtered out, while the right part shows red annotations identifying attributes that undergo imputation.
}   
\label{Fig:MD-AFE}    
\end{figure}

\subsection{Modality Inference Module}
\label{Modality}

\texttt{AutoM\textsuperscript{3}L} begins with the {\bf M}odality {\bf I}nference-LLM (MI-LLM) component, which identifies the associated modality for each column in the structured table. 
To simplify its operation and minimize additional training costs, MI-LLM leverages in-context learning. 
As illustrated in Fig.~\ref{Fig:MD-AFE}(a), the guiding prompt for MI-LLM consists of three essential parts:
(1) An ensemble of curated examples is utilized for in-context learning, assisting MI-LLM in establishing strong correlations between column names and their associated modalities, thereby generating the desired format responses. 
These examples serve as a foundation for MI-LLM to learn from and adapt to the specific dataset at hand, enabling it to accurately infer the modality of each column based on the provided examples.
(2) A subset of the input structured table, consisting of randomly sampled data items paired with their respective column names, is included.
The semantic richness of this subset acts as a guiding force, directing MI-LLM towards accurate identification of modalities. 
By providing a representative sample of the dataset, MI-LLM can better understand the context and characteristics of each column, allowing it to make more informed modality inferences.
(3) User directives go beyond mere instructions, enriching the process with deeper contextual information.
These directives leverage the LLM's exceptional interactivity to enhance the refinement of modality inference.
For instance, a directive such as ``\textit{This dataset delves into the diverse factors influencing animal adoption rates}'' provides MI-LLM with valuable contextual information, enabling a more insightful interpretation of column descriptors.
This additional context helps MI-LLM to make more accurate and relevant modality inferences by considering the overall theme and purpose of the dataset.

\begin{figure}[t]
\centering    
\includegraphics[width=\linewidth]{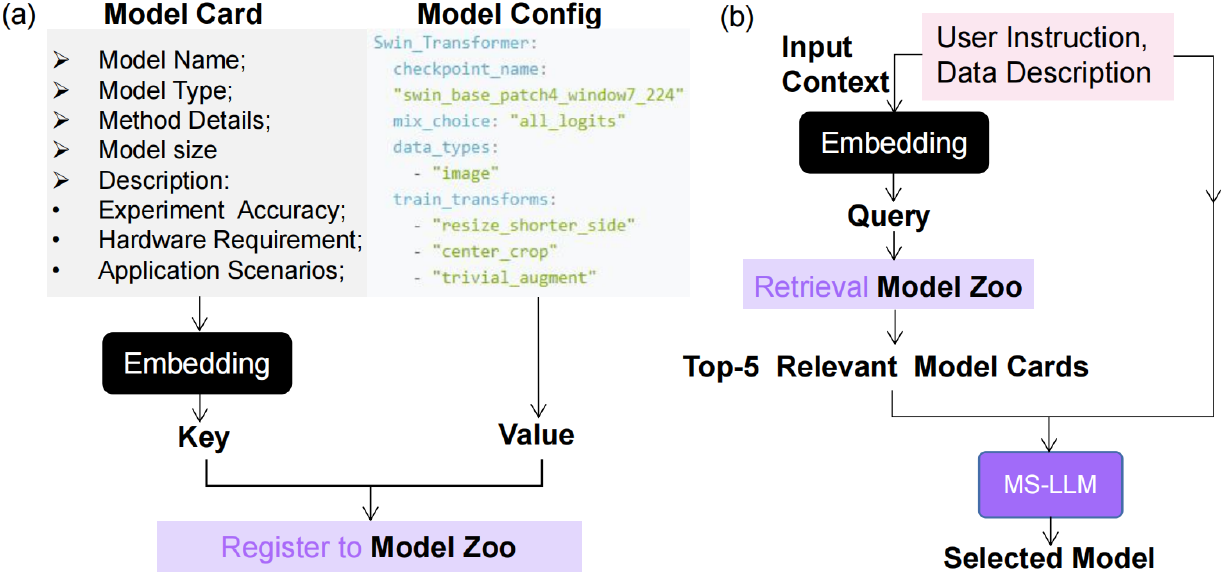}
\caption{Illustration of the model zoo and MS-LLM.
(a) Model addition process: This stage showcases how new models are incorporated into the model zoo, visualized as a vector database. The model card's embedding vector serves as the unique identifier or key, paired with its corresponding model configuration as the value.
(b) Model retrieval process: This stage illustrates the model selection process. Given user directives, the system initiates a query to identify the top 5 models that align with each input modality. From this refined subset, MS-LLM then determines and selects the most suitable model. 
}   
\label{Fig:MS}   
\end{figure}

\subsection{Automated Feature Engineering Module}
Feature engineering is a critical pre-processing phase to address common data challenges, such as handling missing values. 
While conventional AutoML solutions rely heavily on rule-based feature engineering, our \texttt{AutoM\textsuperscript{3}L} framework leverages the exceptional capabilities of LLMs to enhance this process. 
Specifically, we introduce the {\bf A}utomatic {\bf F}eature {\bf E}ngineering-LLM (AFE-LLM), as illustrated in Fig.~\ref{Fig:MD-AFE}(b).
This module utilizes two distinct prompts, resulting in two core components: AFE-LLM\textsubscript{filter} and AFE-LLM\textsubscript{imputed}.
The AFE-LLM\textsubscript{filter} component effectively sifts through the data to eliminate irrelevant or redundant attributes.
On the other hand, the AFE-LLM\textsubscript{imputed} component is dedicated to data imputation, ensuring the completeness and reliability of essential data.
Importantly, these components work together in synergy. After AFE-LLM\textsubscript{filter} refines the features, AFE-LLM\textsubscript{imputed} then addresses relevant data gaps in the dataset.

To enhance feature filtering, AFE-LLM\textsubscript{filter} incorporates the following prompts:
(1) An ensemble of examples for in-context learning, which includes introducing attributes from diverse datasets and intentionally incorporating irrelevant attributes. The objective of AFE-LLM\textsubscript{filter} is to effectively distinguish and eliminate irrelevant attributes
(2) Column names in the structured table, containing abundant semantic information about each feature component, thereby enhancing the LLM's capability to distinguish between crucial and dispensable attributes.
(3) Modality inference results derived from MI-LLM, guiding the LLM to remove attributes of limited informational significance. 
For instance, when comparing a binary attribute that indicates whether someone is over 50 with a continuous attribute such as age, it becomes apparent that the binary attribute may be somewhat redundant. 
In this case, the binary attribute can be identified and removed.
(4) User instructions or task descriptions can be embedded when available, aiming to establish a connection between column names and the corresponding task.

On the other hand, the AFE-LLM\textsubscript{imputed} component is dedicated to data imputation, ensuring the completeness and reliability of essential data. 
Regarding data imputation, AFE-LLM\textsubscript{imputed} exploits its inferential capabilities to effectively identify and fill missing data. 
%
The prompt includes the following:
(1) Data points with missing values, enabling AFE-LLM\textsubscript{imputed} to fill these gaps by discerning patterns and inter-attribute relationships.
(2) A selected subset of data instances from the training set that involves deliberately masking individual attributes and presenting them in Q\&A pairs, laying down an inferential groundwork.
(3) Where available, user instructions or task descriptions are incorporated, offering a richer context and further refining the data imputation process.

Importantly, these components work together in synergy. After AFE-LLM\textsubscript{filter} refines the features, AFE-LLM\textsubscript{imputed} then addresses relevant data gaps in the dataset. By combining feature filtering and data imputation, this module ensures that the dataset is optimized for the subsequent steps in the \texttt{AutoM\textsuperscript{3}L} pipeline.

\subsection{Model Selection Module}
Upon successfully performing the modality inference and automated feature engineering modules, \texttt{AutoM\textsuperscript{3}L} proceeds to determine the optimal model architecture for each data modality. 
The candidate models are cataloged within a model zoo, with each model stored as a model card. 
The model card captures a wide range of details, including the model's name, type, applicable data modalities, empirical performance metrics, hardware requirements, and other relevant information.
To streamline the generation of these cards, we leverage LLM-enhanced tools, such as ChatPaper\cite{ChatPaper}, to eliminate the need for laborious manual writing processes. 
We generate embeddings for these model cards using a text encoder, thereby allowing users to retrieve relevant model cards and seamlessly expand the model zoo by appending new cards, as illustrated in Fig.~\ref{Fig:MS}(a).

Following the model card generation, we propose the {\bf M}odel {\bf S}election-LLM (MS-LLM) to effectively match each modality with the appropriate model.
We view this task as a single-choice dilemma, where the context provides a range of models for selection. 
However, due to limitations on the context length of LLM, it is not feasible to present a complete array of model cards.
Hence, we initially filter the model cards based on their applicable modality type and keep only those that are aligned with the specified data modality. 
Next, a subset of the top 5 models is identified using text-based similarity metrics to compare the user's requirements with the model cards' descriptions.
These high-ranking model cards are then incorporated into the prompt of MS-LLM, along with user instructions and data descriptions.
This combination guides MS-LLM in making its final decision, ultimately identifying the most suitable model for the given modality, as illustrated in Fig.~\ref{Fig:MS}(b).

The MS-LLM prompt fuses the following components:
(1) A selected subset of five model cards, providing insight into potential model candidates.
(2) An input context that intertwines data descriptions and user instructions. The data descriptions clarify important aspects such as data type, label type, and evaluation standards.
Meanwhile, user instructions can provide clarification on specific model requirements. 
For example, a user instruction such as ``deploy the model on the CPU device'' would guide MS-LLM to models optimized for lightweight deployments. This enhances the user-friendliness and intelligence of the framework by enabling interactive execution.

\subsection{Pipeline Assembly Module}
\begin{figure}[t]
\centering    
\includegraphics[width=\linewidth]{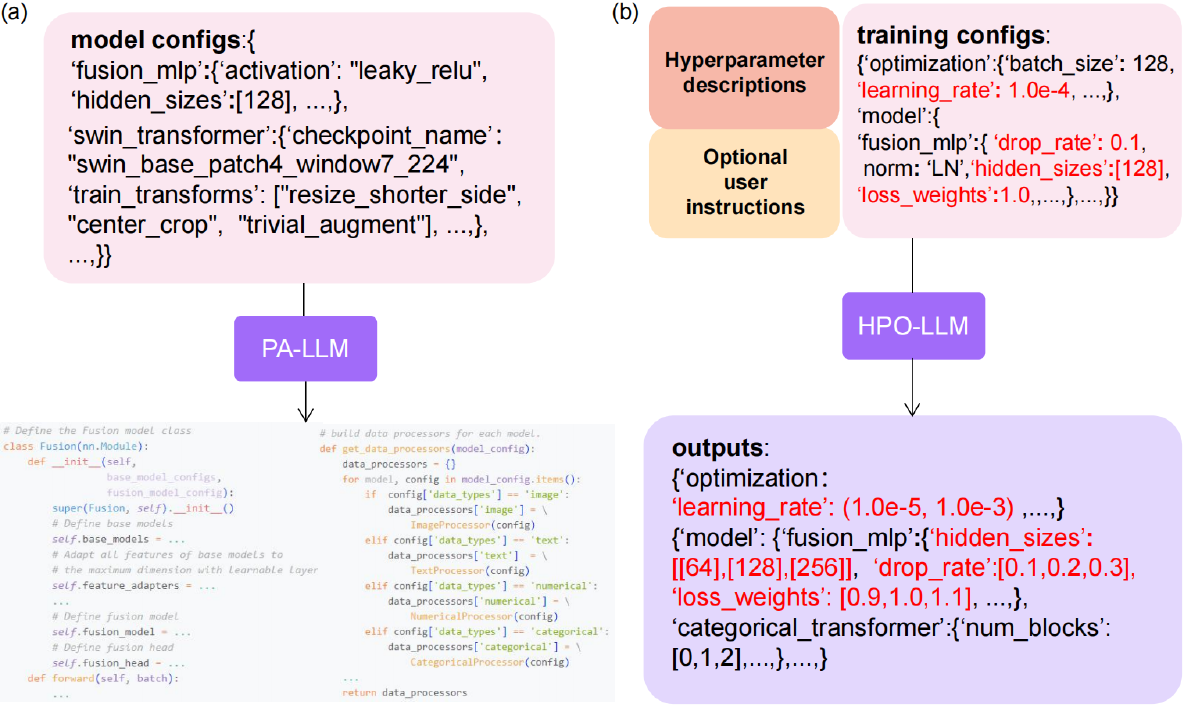}
\caption{
(a) The PA-LLM is responsible for generating executable code, ensuring seamless model training and data processing.
(b) The HPO-LLM deduces hyperparameters and defines search intervals for hyperparameter optimization.
}   
\label{Fig:CG-HPO}    
\end{figure}

After retrieving the unimodal models, a crucial step involves fusing them. We employ a late fusion strategy for integration, which can be mathematically expressed as:
%
%
\begin{equation}
\label{eq1}
\texttt{F}_i = \texttt{feature\_adapter}_i(\texttt{model}_i(\texttt{x}_i)),
\end{equation}
\begin{equation}
\label{eq2}
\texttt{F}_{\textit{cat}} = \texttt{concat}(\texttt{F}_1,...,\texttt{F}_n),
\end{equation}
\begin{equation}
\label{eq3}
\texttt{logits}_{\texttt{fuse}} = 
\texttt{fusion\_head}(\texttt{fusion\_model}(\texttt{F}_{\textit{cat}})),
\end{equation}
where \texttt{concat} denotes concatenation, $\texttt{x}_i$ represents the input data of modality $i$ ($i=1,\cdots,n$), and $\texttt{feature\_adapter}_i$ adapts the output of $\texttt{model}_i$ to a consistent dimension. 
The \texttt{fusion\_head} and \texttt{fusion\_model} are the target models that need to be built.
Determining the architectures for \texttt{fusion\_head} and \texttt{fusion\_model} using rule-based methods that require manual scripting is impractical, as the architectures depend on the number of input modalities.
Instead, we reframe this process as a code generation challenge, where the {\bf P}ipeline {\bf A}ssembly-LLM (PA-LLM) is responsible for generating the fusion model architecture.
PA-LLM leverages the code generation capabilities of LLMs to produce executable code for both model fusion and data processors, as depicted in Fig.~\ref{Fig:CG-HPO}(a).
This is achieved by providing the module with relevant model configuration files within the prompt. 
The data processors are generated based on the specified data preprocessing parameters in the configuration file.
We prioritize the integration of pre-trained models from various modalities, sourced from well-known libraries such as \texttt{HuggingFace} and \texttt{Timm}.
By establishing ties with the wider ML community, we have significantly enhanced the versatility and applicability of our model zoo.

\begin{table*}[t]
\centering
\caption{Task and structure of multimodal datasets}
\label{table_detail}
\begin{tabular}{cccccccc}
\hline
\textbf{Dataset Name} & \textbf{\#Train} & \textbf{\#Test} & \textbf{Task} & \textbf{Metric} & \textbf{Prediction Target} \\ \hline
PAP & 13493 & 1499 & multiclass & accuracy & category of adoption speed \\
MMSD & 17833 & 1981 & binary & auc & whether utterances contains an ironic sentiment\\ 
PPC & 8920 & 993 & regression & rmse & pawpularity score \\
PARA & 25398 & 2822 & regression & rmse& image aesthetics assessment\\      
SPMG & 5000 & 1000 & retrieval & auc & whether data pair is in the same class \\
CH-SIMS & 2052 & 228 & multiclass & accuracy & category of sentiment\\ \hline
\end{tabular}
\end{table*}

\subsection{Automated Hyperparameter Optimization Module}
\label{Automated}
In conventional ML pipelines, hyperparameters such as learning rate, batch size, hidden layer size, and loss weight are commonly adjusted manually, which is labor-intensive and time-consuming. 
Although external tools like \texttt{ray.tune} allow users to conduct optimization by specifying hyperparameters and their search intervals, there is still room for further automation.
To bridge this gap, we propose the {\bf H}yper{\bf P}arameter {\bf O}ptimization-LLM (HPO-LLM), which extends the foundational capabilities of \texttt{ray.tune}.
The core functionality of HPO-LLM lies in its ability to determine optimal hyperparameters and their corresponding search intervals through careful analysis of a provided training configuration file, as depicted in Fig.~\ref{Fig:CG-HPO}(b).
Leveraging the extensive knowledge base of LLMs in ML training, we first utilize LLM to generate comprehensive descriptions for each hyperparameter specified in the configuration file. 
The descriptions, combined with the original configuration file, constitute the prompt for HPO-LLM, which then provides recommendations on the most suitable hyperparameters for optimization.
The input prompt provided to HPO-LLM encompasses the following components:
(1) The training configuration file, containing a comprehensive set of hyperparameters, assists HPO-LLM in selecting the most suitable hyperparameters for optimization.
(2) LLM-generated text descriptions for each hyperparameter, enabling HPO-LLM to gain a comprehensive understanding of the significance of each hyperparameter.
(3) Optional user directives provide a personalized touch, allowing users to incorporate additional instructions that guide HPO-LLM's decision-making process. 
These directives can include emphasizing specific hyperparameters based on unique requirements, resulting in a tailored optimization approach. 
By integrating the capabilities of \texttt{ray.tune} with our HPO-LLM, we have pioneered an approach that enhances hyperparameter optimization by combining automation with advanced decision-making.

\begin{table*}[t]
    \begin{minipage}{0.48\linewidth}
    \centering
    \caption{Evaluation for modality inference. \texttt{AutoM\textsuperscript{3}L} can effectively determine the data modality, even on data that AutoGluon misclassifies or unclassifies. * means the result of manual corrections in modality inference.}
    \label{table_mi}
    \begin{tabular}{llc}
    \toprule
    Dataset & \makebox[1.5cm]{AutoGluon} & \makebox[1.5cm]{AutoM\textsuperscript{3}L} \\ 
    \midrule
    PAP$\uparrow$ & 0.415(0.011) & 0.409(0.014) \\
    MMSD$\uparrow$ & 0.958(0.004) &  0.956(0.004) \\
    PPC$\downarrow$ & 17.78(0.307)   & 17.71(0.315)\\
    PARA$\downarrow$ & 0.568(0.019) & 0.571(0.021)\\
    SPMG$\uparrow$ & 0.985(0.003) & 0.986(0.003)\\
    CH-SIMS$\uparrow$ & $0.543(0.032)^\ast$  & 0.575(0.029)\\
    \bottomrule
    \end{tabular}
    \end{minipage}
    \hfill
    \begin{minipage}{0.48\linewidth}
    \centering
    \caption{Evaluation for feature engineering. \texttt{AutoM\textsuperscript{3}L} filters out noisy features and performs data imputation, effectively mitigating the adverse effects of noisy data. * means the result of manual corrections in modality inference.}
    \label{table_afe}
    \begin{tabular}{lclc}
    \toprule
    Dataset & \makebox[1.22cm]{AutoKeras} & \makebox[1.3cm]{AutoGluon} & \makebox[1.22cm]{AutoM\textsuperscript{3}L} \\ 
    \midrule
    PAP$\uparrow$ & 0.379(0.018) & 0.402(0.014) & 0.407(0.012) \\
    MMSD$\uparrow$& 0.920(0.008) & 0.951(0.004) & 0.956(0.004)\\
    PPC$\downarrow$& 25.18(0.302) & 18.38(0.298) & 17.82(0.304)\\
    PARA$\downarrow$& 0.782(0.025) & 0.576(0.020) & 0.574(0.020)\\
    SPMG$\uparrow$& / &0.984(0.003)& 0.986(0.003)\\
    CH-SIMS$\uparrow$&/ & $0.540(0.031)^\ast$ & 0.575(0.029)\\
    \bottomrule
    \end{tabular}
    \end{minipage}
\end{table*}

\section{Experiments}

\subsection{Experimental Settings}
\subsubsection{Datasets} To assess the effectiveness of the \texttt{AutoM\textsuperscript{3}L} system, we performed experiments on six multimodal datasets, including some obtained from the Kaggle competition platform. 
These datasets cover various tasks, such as classification, regression, and retrieval. 
Table \ref{table_detail} describes the details of the datasets.
%
%
We utilized three classification datasets as follows:
%
(1) PetFinder.my-Adoption Prediction (PAP): This dataset aims to predict the adoptability of pets by analyzing image, text, and tabular modalities.
(2) Multi-Modal Sarcasm Detection (MMSD): This dataset is curated to determine whether an utterance contains ironic sentiment, utilizing image and text modalities.
(3) CH-SIMS: This dataset focuses on sentiment recognition and leverages video and text modalities.
Turning our attention to regression, we utilized two datasets:
(1) PetFinder.my-Pawpularity Contest dataset (PPC): This dataset aims to predict the popularity of shelter pets by leveraging image and tabular modalities.
(2) PARA: This dataset provides diverse image and tabular attributes for personalized image aesthetics assessment.
For the retrieval-based tasks, we employed the Shopee-Price Match Guarantee dataset (SPMG), which aims to determine if two products are identical, relying on data from image and text modalities.
Our performance metrics include accuracy for multiclass classification tasks, the area under the ROC curve (AUC) for binary classification tasks and retrieval tasks, and the root mean square error (RMSE) for regression tasks.
We also evaluated \texttt{AutoM\textsuperscript{3}L} on a large number of unimodal datasets from the AutoML Benchmark\citep{gijsbers2019open} available from OpenML\footnote{\url{https://www.openml.org/}}, which cover regression and binary/multiclass classification tasks.

\subsubsection{Baseline}
Given the scarcity of specialized multimodal AutoML frameworks, our evaluations were exclusively performed using the AutoKeras\footnote{\url{https://github.com/keras-team/autokeras}} and AutoGluon framework. 
AutoKeras is dedicated to neural architecture search (NAS) and hyperparameter optimization for a given dataset.
Setting up training pipelines in AutoGluon required meticulous manual configurations. 
This involved specifying which models to train and conducting an extensive pre-exploration to determine the suitable parameters and their respective search ranges for hyperparameter optimization.
It's crucial to highlight that the automation and intelligence levels of AutoGluon remain challenging to quantify, and in this work, we innovatively measure them through the user study from the human perspective. 
See Appendix \ref{Experiment Implementation} for detailed experimental settings.

\subsubsection{IRB Approval for User Study} 
The user study conducted in this research has received full approval from the Institutional Review Board (IRB).
All methodologies, protocols, and procedures pertaining to human participants were carefully reviewed to ensure they align with ethical standards.

\begin{table*}[t]
\centering
\caption{Evaluation on the hyperparameter optimization. \texttt{AutoM\textsuperscript{3}L}'s self-recommended search space rivals, and in some cases surpasses, manually tuned search spaces. * means the result of manual corrections in modality inference.}
\label{table_result}
\begin{tabular}{lcccccc}
\toprule
Method & PAP$\uparrow$ & MMSD$\uparrow$ & PPC$\downarrow$ & PARA$\downarrow$ & SPMG$\uparrow$ &CH-SIMS$\uparrow$ \\
\midrule
AutoKeras           &    0.385(0.012)       &    0.925(0.007)       &      23.21(0.285)     &   0.744(0.021)        &    /  & /  \\
AutoGluon w/o HPO   &     0.415(0.011)      &    0.958(0.004)       &     17.78(0.307)      &     0.568(0.019)      &   0.985(0.003)  & $0.543(0.032)^\ast$  \\
AutoGluon w/ HPO    &     0.442(0.008)      &    0.963(0.004)       &     17.60(0.217)      &      0.561(0.015)     &    0.990(0.002) & $0.564(0.026)^\ast$  \\
AutoM\textsuperscript{3}L &   0.440(0.012)      &      0.967(0.004)     &    17.47(0.211)       &    0.563(0.016)      &  0.992(0.003) & 0.591(0.027)
    \\
\bottomrule
\end{tabular}
\end{table*}

\subsection{Quantitative Evaluation}

We first carried out quantitative evaluations, drawing direct comparisons with AutoKeras and AutoGluon, focusing on the modality inference, automated feature engineering, and the automated hyperparameter optimization modules.
For modality inference evaluation, apart from the modality inference component, all other aspects of the frameworks are kept consistent.
For feature engineering and hyperparameter optimization, we aligned the modality inference from AutoKeras and AutoGluon with the results of \texttt{AutoM\textsuperscript{3}L} to analyze their respective impacts on performance.
%
To enhance the robustness of our results, we performed 10-fold cross-validation experiments on all datasets. The accuracy is reported as the mean value with its corresponding standard deviation.
Afterwards, we evaluate the pipeline assembly module in terms of intelligence and usability through user study in the next section, due to its inherent difficulty in quantitative evaluation.

\subsubsection{Evaluation for Modality Inference}
Table \ref{table_mi} depicts the comparative performance between AutoGluon's modality inference module and our LLM-based modality inference approach across various multimodal datasets. Since AutoKeras utilizes manually predefined data modality for each column, we excluded it from the comparisons.
Within AutoGluon, modality inference operates based on a set of manually defined rules. For instance, an attribute might be classified as a categorical modality if the count of its unique elements is below a certain threshold. 
Upon observing the results, it's evident that \texttt{AutoM\textsuperscript{3}L} offers accuracy on par with AutoGluon for most datasets. This similarity in performance can be primarily attributed to the congruence in their modality inference outcomes.
However, a notable divergence is observed with the CH-SIMS dataset.
Due to the manually defined rules in AutoGluon being unable to infer video modality and misclassified the "text" attribute as "categorical", the assembly of the training pipeline was hindered, resulting in the failure of the training task.
We manually corrected the misinference of the "text" attribute in AutoGluon, achieving the accuracy of 0.543(0.032). In comparison, \texttt{AutoM\textsuperscript{3}L} demonstrated a significantly superior accuracy, with a notable 3.2\% improvement.
Such a result highlights the robustness of our LLM-based modality inference approach, which effectively infers modality details from column names and associated data through in-context learning with only a few examples, making it significantly more efficient than cumbersome manually designed rules.

\subsubsection{Evaluation for Feature Engineering}
Table \ref{table_afe} illustrates the comparisons of data preprocessing modules using AutoGluon and AutoKeras with our LLM-based automated feature engineering module on multimodal datasets.
Given the completeness of these datasets, we randomly masked portions of the tabular data and manually introduced noisy features from unrelated datasets to assess the effectiveness of automated feature engineering. For datasets without tabular modality, only noise features are introduced. Note that, AutoGluon lacks a dedicated feature engineering module for multimodal data, making this experiment a direct assessment of our automated feature engineering. We observed that automated feature engineering, which implements feature filtering and data imputation, effectively mitigates the impact of noisy data. Across all test datasets, automated feature engineering showed improvements,
while AutoGluon and AutoKeras suffered from performance degradation as they struggled to handle noisy data.
%
%
Since retrieval tasks and video modality inputs are not supported, we did not test AutoKeras on relevant datasets.

\begin{table*}[h]
  \centering
  \caption{Evaluation on single-modal datasets, denoted as $mean(std)^{fails}$(part1). 
  The red values represent the best results.
  }
  \label{table_single1}
  \resizebox{0.78\textwidth}{!}
  {
  \begin{tabular}{clc*{7}{c}}
    \toprule
    \textbf{Task ID} & \textbf{Task Name} & \textbf{Task Type} & \textbf{Task Metric} & \textbf{AUTOGLUON} & \textbf{AUTO-SKLEARN} & \textbf{AUTO-SKLEARN 2} & \textbf{FLAML} & \textbf{GAMA} \\
    \midrule
    146818 & australi... & binary & accuracy & $0.940(0.020)$ & $0.932(0.019)$ & $0.940(0.020)$ & $0.939(0.025)$ & $0.940(0.019)$ \\
    146820 & wilt& binary   & accuracy& $0.994(0.009)$ & $0.994(0.010)$ & $0.995(0.008)$ & $0.988(0.013)$ & $0.996(0.004)$ \\
    167120 & numerai2... & binary  & accuracy& $0.524(0.005)$ & $0.530(0.005)$ & $0.531(0.004)$ & $0.528(0.005)$ & $0.532(0.004)^1$\\
    168757 & credit-g & binary  & accuracy& $0.791(0.039)$ & $0.783(0.042)$ & $0.795(0.038)$ & $0.784(0.039)$ & $0.791(0.030)$\\
    168868 & apsfailu...& binary  & accuracy & $0.992(0.002)$ & $0.992(0.002)$ & $0.992(0.003)$ & $0.992(0.003)$ & $0.992(0.002)$  \\
    190137 & ozone-le... & binary & accuracy & $0.934(0.017)$ & $0.920(0.024)$ & $0.933(0.022)$ & $0.925(0.021)$ & $0.926(0.032)$ \\
    190411 & ada & binary  & accuracy& $0.920(0.018)$ & $0.917(0.017)$ & $0.920(0.018)$ & \textcolor{red}{$0.924(0.018)$} & $0.921(0.018)$ \\
    359955 & blood-tr...& binary   & accuracy& $0.755(0.044)$ & $0.745(0.052)$ & $0.755(0.040)$ & $0.731(0.066)$ & $0.757(0.049)$ \\
    359956 & qsar-bio...& binary  & accuracy & $0.941(0.035)$ & $0.929(0.036)$ & $0.937(0.027)$ & $0.928(0.033)$ & $0.937(0.032)$ \\
    359958 & pc4 & binary  & accuracy& $0.951(0.018)$ & $0.941(0.020)$ & $0.949(0.017)$ & $0.949(0.019)$ & $0.951(0.019)$  \\
    359965 & kr-vs-kp & binary  & accuracy& $1.000(0.000)$ & $1.000(0.000)$ & $1.000(0.000)$ & $1.000(0.000)$ & $1.000(0.000)$ \\
    359930 & quake & regression  & rmse & $0.19(0.0093)$ & $0.19(0.0089)$& - & $0.19(0.0091)$ & $0.19(0.0092)$  \\
    359931 & sensory  & regression  & rmse& \textcolor{red}{$0.67(0.061)$} & $0.69(0.051)$& - & $0.69(0.054)$ & $0.68(0.055)$ \\
    359933 & space ga & regression  & rmse& \textcolor{red}{$0.094(0.013)$} & $0.1(0.025)$& - & $0.1(0.015)$ & $0.096(0.019)$  \\
    359939 & topo21 & regression  & rmse& $0.028(0.0049)$ & $0.028(0.0049)$& - & $0.028(0.0048)$ & $0.028(0.0048)$  \\
    359944 & abalone & regression  & rmse & $2.1(0.12)$ & $2.1(0.11)$ & - &$2.1(0.12)$ & $2.1(0.1)$  \\
    359946 & pol & regression  & rmse& $2.6(0.29)$ & $3.3(0.35)$& - & $3.6(0.37)$ & $3.7(0.3)$  \\
    359936 & elevators & regression  & rmse & \textcolor{red}{$0.0018(5.2e-05)$} & $0.0019(7.3e-05)$& - & $0.002(6.5e-05)$ & $0.0019(6.5e-05)$  \\
    359954 & eucalypt...  & multiclass  & logloss & $0.690(0.053)$ & $0.716(0.047)$ & $0.704(0.061)$ & $0.779(0.121)$ & $0.700(0.057)$  \\
    2073 & yeast & multiclass  & logloss & $1.015(0.087)$ & $1.043(0.080)$ & $1.015(0.084)$ & $1.011(0.083)$ & $1.019(0.081)^5$ \\
    359960 & car & multiclass  & logloss & $0.004(0.011)$ & $0.004(0.008)$ & $0.002(0.004)$ & $0.003(0.005)$ & $0.012(0.008)$ \\
    359964 & dna & multiclass  & logloss & $0.106(0.027)$ & $0.116(0.032)$ & $0.111(0.025)$ & $0.106(0.029)$ & $0.106(0.028)$ \\
    359984 & helena & multiclass  & logloss & \textcolor{red}{$2.470(0.016)$} & $2.526(0.018)$ & $2.485(0.031)$ & $2.564(0.019)$ & $2.731(\text{nan})^9$ \\
    359993 & okcupid-...& multiclass  & logloss& \textcolor{red}{$0.559(0.009)$} & $0.567(0.007)$ & $0.563(0.008)$ & $0.562(0.008)$ & $0.568(0.007)$ \\
    \bottomrule
  \end{tabular}
  }
\end{table*}

\begin{table*}[t!]
\centering
\caption{Evaluation on single-modal datasets, denoted as $mean(std)^{fails}$(part2). 
The red values represent the best results.
}
\label{table_single2}
\resizebox{0.78\textwidth}{!}{
\begin{tabular}{clc*{7}{c}}
    \toprule
    \textbf{Task ID} & \textbf{Task Name}  & \textbf{Task Type} & \textbf{Task Metric} & \textbf{H2O AUTOML} & \textbf{LIGHT AUTOML} & \textbf{MLJAR} & \textbf{TPOT} & \textbf{AUTOM3L} \\
    \midrule
    146818 & australi...  & binary & accuracy& $0.934(0.020)$ & $0.944(0.021)$ & $0.940(0.024)$ & $0.936(0.024)$ & \textcolor{red}{$0.961(0.017)$} \\
    146820 & wilt  & binary & accuracy& $0.993(0.009)$ & $0.994(0.007)$ & $0.994(0.003)^5$ & $0.985(0.025)$  & \textcolor{red}{$0.999(0.002)$}\\
    167120 & numerai2...  & binary & accuracy&$0.531(0.004)$ & $0.531(0.005)$ & $0.530(0.004)$ & $0.527(0.006)$ & \textcolor{red}{$0.534(0.007)$}\\
    168757 & credit-g   & binary & accuracy& $0.782(0.043)$ & $0.788(0.035)$ & - & $0.787(0.034)$  & \textcolor{red}{$0.825(0.032)$} \\
    168868 & apsfailu... & binary & accuracy& $0.992(0.002)$ & \textcolor{red}{$0.994(\text{nan})^9$}  & $0.993(0.002)^6$ & $0.989(0.003)^1$ & $0.990(0.013)$\\
    190137 & ozone-le... & binary & accuracy & $0.930(0.016)$ & $0.930(0.016)$ & $0.911(0.019)^8$ & $0.916(0.026)$  & \textcolor{red}{$0.950(0.015)$} \\
    190411 & ada  & binary & accuracy& $0.921(0.017)$ & $0.922(0.018)$ & $0.921(0.018)$ & $0.917(0.018)$ & $0.920(0.021)$\\
    359955 & blood-tr... & binary & accuracy & $0.760(0.029)$ & 0.749(0.055) & - & $0.754(0.043)$& \textcolor{red}{$0.792(0.045)$}\\
    359956 & qsar-bio...  & binary & accuracy & $0.937(0.037)$ & $0.933(0.033)$ & $0.926(\text{nan})^9$ & $0.933(0.031)$ &\textcolor{red}{$0.949(0.024)$} \\
    359958 & pc4 & binary & accuracy & $0.945(0.022)$ & $0.950(0.016)$ & $0.951(0.017)$ & $0.943(0.023)$ &\textcolor{red}{$0.960(0.017)$}\\
    359965 & kr-vs-kp & binary & accuracy& $1.000(0.000)$ & $1.000(0.000)$ & $1.000(0.000)^7$ & $0.950(0.158)$  & \textcolor{red}{$1.000(0.000)$}\\
    359930 & quake  & regression & rmse & $0.19(0.0094)$ & $0.19(0.0099)$ & $0.19(0.0093)$ & $0.19(0.0096)$ & \textcolor{red}{$0.18(0.0106)$} \\
    359931 & sensory & regression & rmse & $0.7(0.062)$ & $0.69(0.061)$ & $0.67(0.043)$ & $0.68(0.054)$ &$0.7(0.063)$\\
    359933 & space ga & regression & rmse & $0.097(0.012)$ & $0.1(0.017)$ & $0.099(0.018)$ & $0.099(0.018)$ & $0.1(0.011)$ \\
    359939 & topo21  & regression & rmse& $0.028(0.0049)$ & $0.028(0.0049)$ & $0.028(0.0048)$ & $0.028(0.0048)$ & \textcolor{red}{$0.028(0.0042)$}\\
    359944 & abalone  & regression & rmse & $2.1(0.11)$ & $2.1(0.12)$ & $2.1(0.12)$ & $2.1(0.11)$ & \textcolor{red}{$2.1(0.10)$}\\
    359946 & pol & regression & rmse & $3.4(0.28)$ & $3.9(0.33)$ & $2.2(0.23)$ & $3.7(0.38)$ & \textcolor{red}{$2.2(0.16)$}\\
    359936 & elevators  & regression & rmse & $0.002(0.00013)$ & $0.002(5.7e-05)$ & $0.0019(5.8e-05)$ & $0.0019(6.4e-05)$ & $0.0018(6.4e-05)$\\
    359954 & eucalypt...  & multiclass  & logloss & $0.702(0.087)$ & $0.695(0.058)$ & \textcolor{red}{$0.646(0.054)$} & $0.752(0.130)$ & 0.677(0.069)\\
    2073 & yeast  & multiclass  & logloss & $1.040(0.091)$ & $1.038(0.094)^5$ & $1.004(0.085)$ & $1.029(0.083)^5$ & \textcolor{red}{0.995(0.095)}\\
    359960 & car & multiclass  & logloss & $0.001(0.001)$ & $0.002(0.002)$ & $0.002(0.003)$ & $1.450(3.004)$ & \textcolor{red}{0.001(0.001)}\\
    359964 & dna  & multiclass  & logloss& $0.109(0.030)$ & $0.109(0.026)$ & $0.109(0.025)$ & $0.112(0.025)$ & \textcolor{red}{0.098(0.026)}\\
    359984 & helena  & multiclass  & logloss & $2.794(0.018)$ & $2.504(0.014)$ & $2.575(0.021)^1$ & $2.922(0.039)$  & $2.54(0.020)$\\
    359993 & okcupid-... & multiclass  & logloss & $0.567(0.008)$ & $0.560(0.009)$ & $0.563(0.008)$ & $0.569(0.009)$ & $0.565(0.007)$\\
    \bottomrule
    \end{tabular}
    }
\end{table*}

\subsubsection{Evaluation for Hyperparameter Optimization} 
We also conduct experiments to evaluate the automated hyperparameter optimization module within \texttt{AutoM\textsuperscript{3}L}. 
Contrasting with AutoKeras and AutoGluon, which often require users to manually define the hyperparameter search space, \texttt{AutoM\textsuperscript{3}L} simplifies this process.
%
%

From Table \ref{table_result}, it's evident that the integration of hyperparameter optimization during the training phase contributes positively to performance. 
Impressively, \texttt{AutoM\textsuperscript{3}L} matches AutoGluon's accuracy on most datasets and, due to its effective utilization of video information, it has realized a 2.7\% improvement on the CH-SIMS dataset.
However, the standout advantage of \texttt{AutoM\textsuperscript{3}L} lies in its automation. 
%
While AutoGluon requires a manual setup, which can often be tedious, \texttt{AutoM\textsuperscript{3}L} significantly reduces the need for human intervention, providing a more seamless and automated experience.
Another finding is that AutoKeras achieves lower accuracy on all datasets. In our analysis, we attribute it to the network structures obtained within its limited network search space, which lacks pretraining on large-scale datasets.
In contrast, our approach leverages the strength of pretrained models by linking with open-source communities such as HuggingFace and Timm. 
This integration allows us to access more powerful pretrained models, contributing to the improved performance demonstrated in our work.



\subsubsection{Uni-Modal Scenario Evaluation}
Given that most AutoML frameworks currently focus on single-modality AutoML, to demonstrate the scalability of \texttt{AutoM\textsuperscript{3}L}, we also evaluated \texttt{AutoM\textsuperscript{3}L} on a large-scale tubular modality AutoML Benchmark\cite{gijsbers2019open} from OpenML. 
%
%
We compared it with a plethora of popular single-modal AutoML frameworks\cite{erickson2020autogluon,vakhrushev2021lightautoml,wang2021flaml,ledell2020h2o,olson2016tpot,feurer2020auto,gijsbers2021gama} on large and representative  datasets 
covering binary classification, multi-class classification, and regression tasks.
The metrics include logarithmic loss for multi-class classification tasks, the root mean squared error for regression tasks, and the area under the ROC curve for binary classification tasks. We reported the mean and standard deviation based on 10-fold cross-validation.
In the experiment, we employed the user instruction: ``\textit{the model with the best performance on tabular data}'' to drive MS-LLM for model selection, and FT-Transformer was chosen and integrated with other components to form a training pipeline.
The results in Table \ref{table_single1} and Table \ref{table_single2} demonstrate \texttt{AutoM\textsuperscript{3}L}'s strong performance even in single-modality settings.
%
%
Notably, in most frameworks compared, model ensemble techniques are employed to produce final predictions. However, \texttt{AutoM\textsuperscript{3}L} solely utilized a single model for evaluation and achieved competitive results, even outperforming others on most experimental tasks. 

%

\begin{figure}[t]
    \centering
    \includegraphics[width=1.02\linewidth]{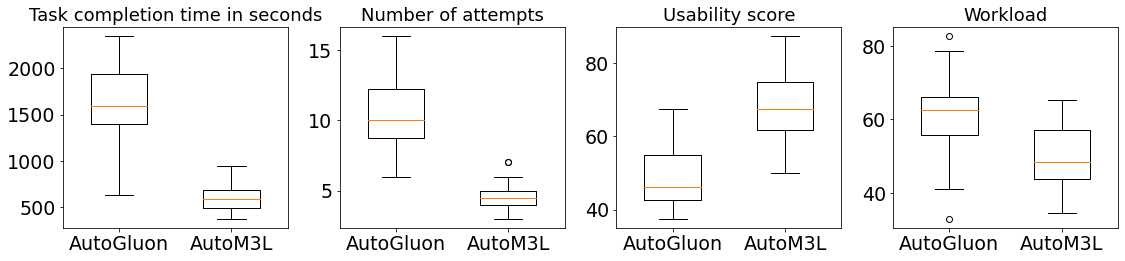}
    \caption{Boxplots displaying the distribution of the four variables collected in the user study.}
    \label{boxplots}
\end{figure}

\begin{figure}[t]
    \centering    
    \includegraphics[width=\linewidth]{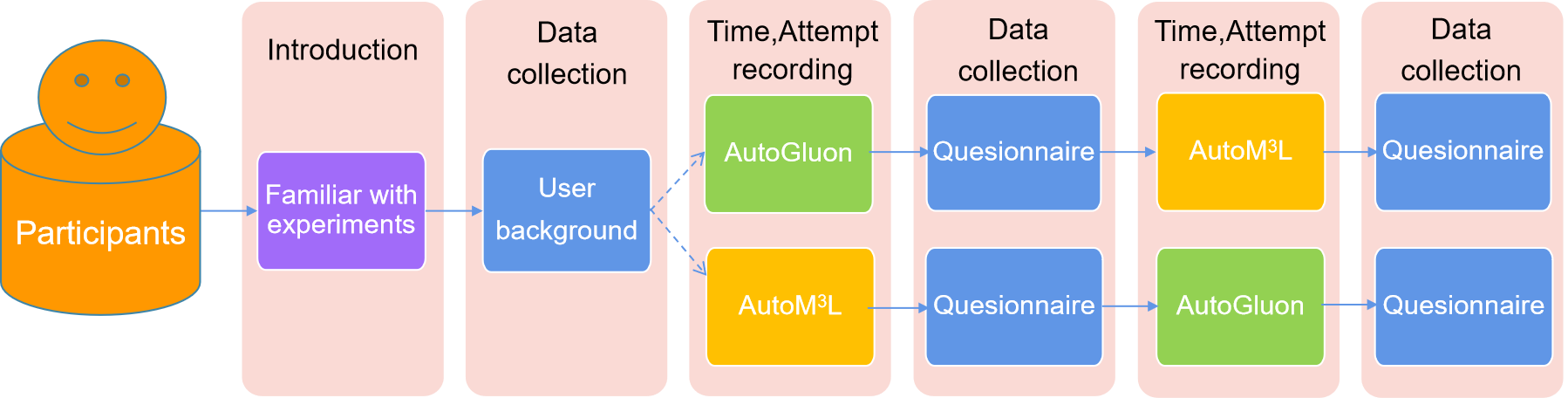}
    \caption{The workflow of the user study to measure the user-friendliness of the \texttt{AutoM\textsuperscript{3}L}.}
    \label{userstudy}    
\end{figure}

\subsection{User Study}

\subsubsection{Hypothesis Formulation and Testing}
To assess \texttt{AutoM\textsuperscript{3}L}'s effectiveness, we conducted a user study focused on whether the LLM controller can enhance the degree of automation within the multimodal AutoML framework.
We formulated null hypotheses:
\begin{itemize}
\item \textbf{H1}: \textit{\texttt{AutoM\textsuperscript{3}L} does \textbf{not} reduce time required for learning and using the framework.}
\item \textbf{H2}: \textit{\texttt{AutoM\textsuperscript{3}L} does \textbf{not} improve user action accuracy.}
\item \textbf{H3}: \textit{\texttt{AutoM\textsuperscript{3}L} does \textbf{not} enhance overall framework usability.}
\item \textbf{H4}: \textit{\texttt{AutoM\textsuperscript{3}L} does \textbf{not} decrease user workload.}
\end{itemize}

We performed single-sided t-tests to evaluate statistical significance. Specifically, we compared \texttt{AutoM\textsuperscript{3}L} and AutoGluon on the following variables: task execution time, the number of attempts, system usability, and perceived workload. 
%

\subsubsection{User Study Design} 
As depicted in Fig.~\ref{userstudy}, our user study's workflow unfolds in structured phases. 
Note that the user study has been reviewed by IRB and granted full approval.
The study begins with the orientation phase where voluntary participants are acquainted with the objectives, underlying motivations, and procedural details of the user study. 
This phase is followed by a user background survey, which gleans insights into participants' professional roles, their prior exposure to technologies such as LLM and AutoML, and other pertinent details.
The core segment of the study involves hands-on tasks that participants undertake in two distinct conditions: perform multimodal task AutoML with AutoGluon and with \texttt{AutoM\textsuperscript{3}L}.
These tasks center around exploring the automation capabilities of the AutoML frameworks, as well as gauging the user-friendliness of their features. 
Participants, guided by clear instructions, are tasked with constructing multimodal training pipelines employing certain models and defining specific hyperparameter optimization domains.

To ensure a balanced perspective, participants are randomly split into two groups: the first interacts with AutoGluon, while the second delves into \texttt{AutoM\textsuperscript{3}L}.
Upon task completion, the groups swap platforms. 
For a holistic understanding of user interactions, we meticulously track both the time taken by each participant for task execution and the number of attempts before the successful execution.
The study culminates with a feedback session, where participants articulate their impressions regarding the usability and perceived workload of both AutoGluon and \texttt{AutoM\textsuperscript{3}L} via questionnaire. 
Their feedback to the questionnaire, captured using Google Forms, form a crucial dataset for the subsequent hypothesis testing and analysis.
Our study cohort consisted of 20 diverse participants:
6 software developers, 10 AI researchers, and 4 students, which ensured a rich blend of perspectives of the involved users. 

\begin{table}[t]
\caption{Hypothesis testing results from paired two-sample one-sided t-tests.
}
\label{Hypothesis_testing}
\centering
\resizebox{1\linewidth}{!}{
\begin{tabular}{cccc}
\toprule
\textbf{Hypothesis} & \textbf{T Test Statistic} & \textbf{P-value} & \textbf{\begin{tabular}[c]{@{}c@{}}Null Hypothesis\end{tabular}} \\
\midrule
\textbf{H1}         & 12.321                  & $8.2 \times {10}^{-11}$                 & Reject                                                                 \\
\textbf{H2}         & 10.655                   & $9.3 \times {10}^{-10}$                & Reject                                                                  \\
\textbf{H3}         & -5.780                  &    $1.0 \times {10}^{-5}$             & Reject                                                                  \\
\textbf{H4}         & 3.949                   &   $4.3 \times {10}^{-4}$             & Reject                                                                  \\
\bottomrule
\end{tabular}
}
\end{table}

\subsubsection{Results and Analysis of Hypothesis Testing} 
The data we gathered spanned four variables, visualized in Fig.~\ref{boxplots}.
To validate our hypotheses, we performed paired two-sample t-tests (essentially one-sample, one-sided t-tests on differences) for the aforementioned variables across two experimental conditions: AutoGluon and \texttt{AutoM\textsuperscript{3}L}.
These tests were conducted at a significance level of 5\%. 
The outcomes in Table \ref{Hypothesis_testing} empower us to reject all the null hypotheses, underscoring the superior efficacy and user-friendliness of \texttt{AutoM\textsuperscript{3}L}.
The success of \texttt{AutoM\textsuperscript{3}L} can be largely attributed to the interactive capabilities endowed by LLMs, which significantly reduce the learning curve and usage costs.
%

Since most researchers were familiar with LLMs but had limited AutoML experience, increasing their learning curve on AutoGluon. Whereas the majority of engineers and students were novices in both these spheres, facing steeper challenges in grasping AutoGluon. Interestingly, even researchers acquainted with AutoML felt that \texttt{AutoM\textsuperscript{3}L} demonstrated superior ease of use comparatively. Collectively across backgrounds, \texttt{AutoM\textsuperscript{3}L} attained higher user ratings, lower task completion times, and fewer failed attempts, which quantitatively validates its improved user-friendliness.

\section{Conclusion}
In this work, we introduce \texttt{AutoM\textsuperscript{3}L}, an LLM-powered Automated Multimodal Machine Learning framework.
\texttt{AutoM\textsuperscript{3}L} explores automated pipeline construction, automated feature engineering, and automated hyperparameter optimization. 
This enables the realization of an end-to-end multimodal AutoML framework.
Leveraging the exceptional capabilities of LLMs, \texttt{AutoM\textsuperscript{3}L} provides adaptable and accessible solutions for multimodal data tasks.
It offers automation, interactivity, and user customization. 
Through extensive experiments and user studies, we demonstrate \texttt{AutoM\textsuperscript{3}L}'s effectiveness and user-friendliness. This highlights its potential to transform multimodal AutoML.
\texttt{AutoM\textsuperscript{3}L} marks a significant advance, offering enhanced multimodal machine learning across domains. 
Our future direction is to encompass a diverse range of data modalities, spanning graph, audio, and point clouds, among others.
%


\bibliographystyle{ACM-Reference-Format}
\bibliography{sample-base}


\clearpage
\appendix
\section*{Appendix} 
\addcontentsline{toc}{section}{Appendix}

In this appendix, we provide additional details of our approach. We present complete prompts for each LLM module in Sec.A,
while Sec.B offers comprehensive details about the datasets.
In Sec.C and Sec.D, we provide more details about user study, including evaluation metrics, participant background, etc. 
More detailed experimental implementation and related work are given in Sec.E and Sec.F. 
Finally, we discuss the impact of the bias problem of LLMs, computation cost, etc. of our framework in Sec.G.

\section{Prompts}
\label{appendix-prompts}

\begin{samepage}
\begin{center}
\textbf{Prompt 1: Full System Prompt For MI-LLM.} 
\end{center}
\begin{tcolorbox}[colback=gray!10!white,colframe=gray!40!black]
You are a helpful assistant that analyzes data modalities in multimodal Auto-Machine learning task.

Your task is to analyze the data type of each column of the pandas.DataFrame tabular data.

Your answer must be in a strict JSON format: \{``column name'': ``data type''\}.

You can analyze the data type based on the corresponding column name,column data and the user instructions, which may include the context of tasks/datasets, etc..

You should not omit any column of data in your answer.
\vspace{2mm}

Here are some examples for your reference:

\textcolor{blue}{Input: instructions:\{case1\_desc\},Date:\{case1\_input\}}

\textcolor{blue}{Output: \{case1\_output\}}

\textcolor{blue}{...}





\textcolor{red}{Input: instructions:\{data\_desc\},Date:\{data\_input\}}

\textcolor{red}{Output:}
\end{tcolorbox}
\end{samepage}

\begin{samepage}
\begin{center}
\textbf{Prompt 2: Full System Prompt For AFE-LLM\textsubscript{filter}.} 
\end{center}
\begin{tcolorbox}[colback=gray!10!white,colframe=gray!40!black]
You are a helpful assistant that applies feature engineering, especially feature selection.

Given a set of features, your task is to filter out some features that are not relevant to the specific task.

You should filter out the features based on the feature names, feature type and user instrucions, which may contain the context of tasks/datasets, etc..

You cannot forge features that are not in the Input.

In particular, image features should be preserved.

\vspace{12pt}
Here are some examples for your reference:

\textcolor{blue}{Input: instructions:\{case1\_task\}, features type:\\\{case1\_feature\_type\}, features:\{case1\_feature\}}

\textcolor{blue}{Output: \{case1\_retained\_feature\}}

\textcolor{blue}{...}


\textcolor{red}{Input: instructions:\{data\_task\}, features type\\:\{data\_feature\_type\}, features:\{data\_feature\}}

\textcolor{red}{Output:}

\end{tcolorbox}
\end{samepage}

\begin{samepage}
\begin{center}
\textbf{Prompt 3: Full System Prompt For AFE-LLM\textsubscript{imputed}.} 
\end{center}
\begin{tcolorbox}[colback=gray!10!white,colframe=gray!40!black]
You are a helpful assistant that applies feature engineering, especially data imputation.

Given a feature sequence, your task is to predict missing values in it. Missing values are represented by "???".

You should predict missing values based on other feature values in the sequence and, you can refer to user instructions, which may contrain context of the task/dataset, etc...

Your output format must be a certain element value, don't reply the reasoning process.

\vspace{12pt}
Here are some examples for your reference:

\textcolor{blue}{Input: instructions:\{case1\_task\}, feature sequence:\\\{case1\_data\_sequence\}}

\textcolor{blue}{Output: \{case1\_miss\_value\}}

\textcolor{blue}{...}





\textcolor{red}{Input: instructions:\{data0\_task\}, feature sequence:\\\{data0\_sequence\}}

\textcolor{red}{Output:}

\end{tcolorbox}
\end{samepage}

\begin{samepage}
\begin{center}
\textbf{Prompt 4: Full System Prompt For HPO-LLM.} 
\end{center}
\begin{tcolorbox}[colback=gray!10!white,colframe=gray!40!black]

You are a helpful assistant that infers the hyperparameters and their search ranges for hyperparameter optimization in machine learning task.

You can use the format:[,,,] to represent a discrete search range.

You can choose up 3 hyperparameters that you think are most suitable for hyperparameter optimization. 

Your answer must be in a strict JSON format: \{``hyperparameter\_name'':``search\_range''\}.





\vspace{8pt}
Here are some things you need to focus on:

(1).If the values in the search space are of type INT or FLOAT, then the search space needs to have at least 3 values.

(2).The search ranges should refer to the original value of the config. The search ranges should include the original value of the config.

(3).You should not output the hyperparameters don't need to optimize.

(4).You cannot forge parameters that are not in the configuration file.

(5).If the ``checkpoint\_name'' is in config, only the ``loss\_weight'' is taken.

\vspace{8pt}

\textcolor{red}{Here are some comments to help you understand the parameters better: \{self\_desc\}}

\textcolor{red}{Given the config as follow: \{config\}}

\textcolor{red}{Given the user requirements: \{user requirements\}}

\textcolor{red}{Your answer:}

\end{tcolorbox}
\end{samepage}

\begin{samepage}
\begin{center}
\textbf{Prompt 4: Full System Prompt For MS-LLM.} 
\end{center}
\begin{tcolorbox}[colback=gray!10!white,colframe=gray!40!black]
I am a deep learning software develop engineer, you're a code compiler, and we're working together on a multimodal Auto-Machine learning task.

Given the dataset description and user request , your task is to help the user to select a suitable model.

You should focus more on the description of the models and find the model that has the most potential to solve requests and tasks.

Your answer must be in a strict JSON format: \{``name'': ``model name'', ``reason'': ``your reasons to select the model''\}.
Please choose the most suitable model from:
\{model\_cards\}

\vspace{12pt}
\textcolor{red}{User: Assume we have a dataset:\{data\_desc\} and user request: \{user\_request\},please select the most suitable model.}

\textcolor{red}{Your answer:}

\end{tcolorbox}
\end{samepage}

\begin{samepage}
\begin{center}
\textbf{Prompt 5: Full System Prompt For PA-LLM(Data Processors Generation).} 
\end{center}
\begin{tcolorbox}
You are a helpful assistant that writes data processors code to load different types of data for multimodal Auto-Machine learning task.

Since different types of models need different data preprocessing, your task is to write a function to return the corresponding data processors based on models' config.

Specifically, you do not need to define the data processor for fusion model, and the label data processor is also required to provide label data for each model.

The function return must be in a strict dict format: \{``data type'': ``data processor''\}.

Please specify the library you imported in the code.

\vspace{4pt}
Here are some data processors code for you reference:

\vspace{4pt}
from multimodal.data import ImageProcessor

class ImageProcessor:

\quad\quad
def \_\_init\_\_(self,model\_config):

\quad\quad    
...
    
\vspace{4pt}
from multimodal.data import TextProcessor

class TextProcessor:

\quad\quad 
def \_\_init\_\_(self,model\_config):

\quad\quad
    ...
    
\vspace{4pt}
from multimodal.data import CategoricalProcessor

class CategoricalProcessor:

\quad\quad
def \_\_init\_\_(self,model\_config):

\quad\quad    
...
    



% ...
    



% ...

...
    
\vspace{4pt}
\textcolor{red}{Given some models' config as follow:\{configs\}}

\textcolor{red}{Your answer:}

\end{tcolorbox}
\end{samepage}

\begin{samepage}
\begin{center}
\textbf{Prompt 6: Full System Prompt For PA-LLM(Pipeline Assembly).} 
\end{center}
\begin{tcolorbox}
\small
You are a helpful assistant that writes the Deep learning model code.
You task is to write a fusion model to fuse different base models' features.
Use \# before every line except the python code.
Here are some model code for you reference:

from multimodal.models import CategoricalTransformer

class CategoricalTransformer(nn.Module):

\quad\quad
def \_\_init\_\_(self,model\_config):

\quad\quad    
...
    
from multimodal.models import NumericalTransformer

class NumericalTransformer(nn.Module):

\quad\quad 
def \_\_init\_\_(self,model\_config):

\quad\quad
    ...
    
from multimodal.models import TimmAutoModelForImagePrediction

class TimmAutoModelForImagePrediction(nn.Module):

\quad\quad
def \_\_init\_\_(self,model\_config):

\quad\quad    
...
    
from multimodal.models import HFAutoModelForTextPrediction

class HFAutoModelForTextPrediction(nn.Module):

\quad\quad
def \_\_init\_\_(self,model\_config):

\quad\quad    
...

Given some base models' config as follow:\{base\_configs\};
Give the fusion model config as follow:
\{fusion\_config\}

You should then respond to me the code with:

1). Fusion technique should be learnable, MLP is recommended.

2). The fusion model structure should be defined as fusion\_model and fusion\_head,which output features and logits, respectively.

3). Base models instance should be defined in Fusion model Class.You should not change the value of the output of base model instances.

4). All base models have a uniform variable(self.out\_features\_dim) to represent the output features dimension.

5). Finding the maximum dimension of all base models' output features, and define learnable linear layers to adapt all base models' output features to the maximum dimension as the input of fusion\_model. For example, if three models have feature dimensions are [512, 768, 64], it will linearly map all the features to dimension 768.

6). Output the logits,features,loss weights of fusion model and base models.The return must be in a JSON format: \{model\_name:\{``logits'':...,``features'':...,``weight'':...\}\}.

7). All the network layers and variable self.model\_name,self.loss\_weight should be defined in function \_\_init\_\_, not in function forward.

8). Some variables are not present in each model's config,you cannot use a variable that does not exist in the corresponding model config.


You should only respond in the format as described below :

Class Fusion:

\quad\quad
    def \_\_init\_\_(self,...)

\quad\quad
    ...

\quad\quad
    def forward(self,batch)

\quad\quad
    ...
    
\quad\quad
    fusion\_features = self.fusion\_model(...)

\quad\quad
    fusion\_logits   = self.fusion\_head(fusion\_features)

\quad\quad
    ...

\end{tcolorbox}
\end{samepage}








\section{Structured Table Datasets}
\label{EXAMPLES_OF_STRUCTURED_TABLE}

For the purpose of reproducibility, we provide the downloading links to the datasets used in this work and, Table \ref{example1} and Table \ref{example2} describe the details of the sample datasets.
\begin{itemize}
\item 
 PetFinder.my-Adoption Prediction dataset (PAP):
 
 \url{https://www.kaggle.com/competitions/petfinder-adoption-prediction}

\item 
PetFinder.my-Pawpularity Contest dataset (PPC)

\url{https://www.kaggle.com/competitions/petfinder-pawpularity-score}

\item 
Multi-Modal Sarcasm Detection (MMSD):

\url{https://github.com/headacheboy/data-of-multimodal-sarcasm-detection}

\item 
Shopee-Price Match Guarantee dataset (SPMG):

\url{https://www.kaggle.com/competitions/shopee-product-matching}

\item 
PARA:

\url{https://cv-datasets.institutecv.com/#/data-sets}

\item 
CH-SIMS:

\url{https://github.com/thuiar/MMSA}

\end{itemize}


\begin{table*}[t]
\centering
\caption{Example of data in multimodal structured table dataset with text (name, description), numerical (age), categorical (gender), and image paths (images) columns. With these attributes, we want to predict how quickly the pet will be adopted (adoption\_speed). We only display the partial columns for brevity.}
\label{example1}
\begin{tabular}{ccclcc}
\hline
\cellcolor[HTML]{FFFFFF}\textbf{name} & \multicolumn{1}{r}{\cellcolor[HTML]{FFFFFF}\textbf{age}} & \multicolumn{1}{r}{\cellcolor[HTML]{FFFFFF}\textbf{gender}} & \multicolumn{1}{c}{\cellcolor[HTML]{FFFFFF}\textbf{description}}                                 & \textbf{images}        & \multicolumn{1}{l}{\textbf{adoption}} \\ \hline
\rowcolor[HTML]{DFDFDF} 
Coco                                  & 13                                                       & 2                                                           & \begin{tabular}[c]{@{}l@{}}Hi, Coco is a \\ rescued puppy from \\ xthe streets, ...\end{tabular} & images/640683dd9-1.jpg & 0                                     \\
Muffin                                & 1                                                        & 2                                                           & \begin{tabular}[c]{@{}l@{}}This is the puppy\\  we adopted from \\ Crystal, ...\end{tabular}     & images/e3935c62d-1.jpg & 0                                     \\
\rowcolor[HTML]{DFDFDF} 
Usyang                                & 4                                                        & 1                                                           & \begin{tabular}[c]{@{}l@{}}Both of my kitten\\  is so active and \\ spoilt, ...\end{tabular}     & images/d33f713d0-1.jpg & 1                                     \\
...                                   & ...                                                      & ...                                                         & \multicolumn{1}{c}{...}                                                                          & ...                    & ...                                   \\ \hline
\end{tabular}
\end{table*}

\begin{table*}[t]
\centering
\caption{Example of data in multimodal structured table dataset with categorical attribute (eyes, face, near, blur) and corresponding photo paths (images) of pets. With these attributes, we want to determine a pet photo’s appeal (pawpularity). We only display the partial columns for brevity.}
\label{example2}


\begin{tabular}{cccccc}
\hline
\rowcolor[HTML]{FFFFFF} 
\textbf{eyes}            & \textbf{face}            & \textbf{near}            & \textbf{blur}            & \textbf{images}                                                                                                      & \textbf{pawpularity} \\ \hline
\rowcolor[HTML]{DFDFDF} 
{\color[HTML]{333333} 1} & {\color[HTML]{333333} 1} & {\color[HTML]{333333} 1} & {\color[HTML]{333333} 0} & {\color[HTML]{333333} \begin{tabular}[c]{@{}c@{}}train\_images/\\ 0007de18844b0dbbb5e1f607da0606e0.jpg\end{tabular}} & 63                   \\
1                        & 1                        & 0                        & 0                        & \begin{tabular}[c]{@{}c@{}}train\_images/\\ 0009c66b9439883ba2750fb825e1d7db.jpg\end{tabular}                        & 42                   \\
\rowcolor[HTML]{DFDFDF} 
1                        & 1                        & 1                        & 0                        & \begin{tabular}[c]{@{}c@{}}train\_images/\\ 0013fd999caf9a3efe1352ca1b0d937e.jpg\end{tabular}                        & 28                   \\
...                      & ...                      & ...                      & ...                      & ...                                                                                                                  & ...                  \\ \hline
\end{tabular}
\end{table*}

\begin{table*}[t]
\label{example3}
\centering
\caption{Example of data in multimodal structured table dataset with image paths (image1, image2) and texts (title1, title2). We want to determine whether the image-text and image-text pair is in same class(p=1) or not. The original data give a image path, it's text description and corresponding class.}

\begin{tabular}{ccccc}
\hline
\rowcolor[HTML]{FFFFFF} 
\textbf{image1}                                                                                             & \textbf{title1}                                                                                                     & \textbf{image2}                                                                                             & \textbf{title2}                                                                                  & \textbf{p}               \\ \hline
\rowcolor[HTML]{DFDFDF} 
{\color[HTML]{333333} \begin{tabular}[c]{@{}c@{}}f28094791c585c3f\\ 1f7c0662e2cbecee\\ .jpg\end{tabular}} & {\color[HTML]{333333} \begin{tabular}[c]{@{}c@{}}YANG YY 001 \\ Air pump \\ aerator baterai \\ Yang\end{tabular}} & {\color[HTML]{333333} \begin{tabular}[c]{@{}c@{}}a4e379e2da3947ce\\ d71630fbdda70c4b\\ .jpg\end{tabular}} & {\color[HTML]{333333} \begin{tabular}[c]{@{}c@{}}Paket Super \\ Kinclong Lengkap\end{tabular}} & {\color[HTML]{333333} 0} \\
\begin{tabular}[c]{@{}c@{}}1267eb326c6ad70a\\ 32fb942b4834f818\\ .jpg\end{tabular}                        & \begin{tabular}[c]{@{}c@{}}Promag Tablet \\ 1 Box\end{tabular}                                                    & \begin{tabular}[c]{@{}c@{}}2d8ca235317a263c\\ aeb5432e57aeeff8\\ .jpg\end{tabular}                        & \begin{tabular}[c]{@{}c@{}}Promag 1 Box isi \\ 3 lembar\end{tabular}                           & 1                        \\
\rowcolor[HTML]{DFDFDF} 
\begin{tabular}[c]{@{}c@{}}2d8ca235317a263c\\ aeb5432e57aeeff8\\ .jpg\end{tabular}                        & \begin{tabular}[c]{@{}c@{}}Promag 1 Box isi \\ 3 lembar\end{tabular}                                              & \begin{tabular}[c]{@{}c@{}}088fec7809a7d809\\ 73606507b123c66d\\ .jpg\end{tabular}                        & \begin{tabular}[c]{@{}c@{}}PAKET SHAMPO \\ KUNTZE\end{tabular}                                 & 0                        \\
...                                                                                                       & ...                                                                                                               & ...                                                                                                       & ...                                                                                            & ...                      \\ \hline
\end{tabular}
\end{table*}

\section{Questionnaire and Variables in User Study}
\label{Questionnaire and Variables in User Study}
\subsection{User Background Survey Questionnaire}
\begin{enumerate}[leftmargin=*]

    \item Age? \textit{Single-choice question.}
    \begin{enumerate}
        \item[$\bigcirc$] \textless18
        \item[$\bigcirc$] 18-24
        \item[$\bigcirc$] 25-34
        \item[$\bigcirc$] 35-44
        \item[$\bigcirc$] \textgreater44
    \end{enumerate}

    \item Gender? \textit{Single-choice question.}
    \begin{enumerate}
        \item[$\bigcirc$] Male
        \item[$\bigcirc$] Female
    \end{enumerate}

    \item What is your highest level of education? \textit{Single-choice question.}
    \begin{enumerate}
        \item[$\bigcirc$] High School or Below
        \item[$\bigcirc$] Bachelor's Degree
        \item[$\bigcirc$] Master's Degree
        \item[$\bigcirc$] Ph.D.
        \item[$\bigcirc$] Other: \underline{\qquad\qquad\qquad\qquad\qquad\qquad\qquad}
    \end{enumerate}

    \item What is your occupation? \textit{Single-choice question.}
    \begin{enumerate}
        \item[$\bigcirc$] Student
        \item[$\bigcirc$] Engineer
        \item[$\bigcirc$] Data Scientist/Analyst
        \item[$\bigcirc$] AI Algorithm Engineer
        \item[$\bigcirc$] Educator
        \item[$\bigcirc$] Doctor/Medical Professional
        \item[$\bigcirc$] Other: \underline{\qquad\qquad\qquad\qquad\qquad\qquad\qquad}
    \end{enumerate}

    \item Are you familiar with Python? \textit{Single-choice question.}
    \begin{enumerate}
        \item[$\bigcirc$] Yes
        \item[$\bigcirc$] No
    \end{enumerate}

    \item Are you familiar with terminal operation? \textit{Select only one bullet point.}
    \begin{enumerate}
        \item[$\bigcirc$] Yes
        \item[$\bigcirc$] No
    \end{enumerate}

    \item Do you have any experience with machine learning? \textit{Select only one bullet point.}
    \begin{enumerate}
        \item[$\bigcirc$] Yes, experienced
        \item[$\bigcirc$] Yes, some experience
        \item[$\bigcirc$] No, no experience
    \end{enumerate}

    \item Have you used any AutoML tools or platforms before? \textit{Select only one bullet point.}
    \begin{enumerate}
        \item[$\bigcirc$] Yes, very familiar
        \item[$\bigcirc$] Yes, somewhat familiar
        \item[$\bigcirc$] No, not familiar
    \end{enumerate}

    \item Are you familiar with the AutoGluon used in this experiment? \textit{Select only one bullet point.}
    \begin{enumerate}
        \item[$\bigcirc$] Yes
        \item[$\bigcirc$] No
    \end{enumerate}

    \item Are you familiar with the Large language model? \textit{Select only one bullet point.}
    \begin{enumerate}
        \item[$\bigcirc$] Yes, very familiar
        \item[$\bigcirc$] Yes, somewhat familiar
        \item[$\bigcirc$] No, not familiar
    \end{enumerate}

    \item Would you be willing to participate in this experiment? \textit{Select only one bullet point.}
    \begin{enumerate}
        \item[$\bigcirc$] Yes, I am willing to participate
        \item[$\bigcirc$] No, I am not willing to participate
    \end{enumerate}

    \item What are your expectations for automated machine learning methods? \textit{Select only one bullet point.}
    \begin{enumerate}
    \item \underline{\qquad\qquad\qquad\qquad\qquad\qquad\qquad\qquad\qquad}
    \end{enumerate}
\end{enumerate}

\subsection{Questionnaire After Task Execution}
\label{Questionnaire After Task Execution}
\begin{enumerate}[leftmargin=*]


    \item How much time did it take in total to complete all the tasks? (in seconds) \underline{\qquad\qquad}

    \item How many script execution attempts did you make in total to complete the tasks?\underline{\qquad\qquad}

    \item I think that I would like to use this system frequently.
    \textit{Select only one bullet point.}\\
    \begin{tabular}{rcccl}
    1 & 2 & 3 & 4 & 5\\
    \hline
     Strongly Disagree  $\bigcirc$  & $\bigcirc$ &  $\bigcirc$  &  $\bigcirc$  &  $\bigcirc$  Strongly Agree \\
    \hline
    \end{tabular}

    \item I found the system unnecessarily complex.
    \textit{Select only one bullet point.}\\
    \begin{tabular}{rcccl}
    1 & 2 & 3 & 4 & 5\\
    \hline
     Strongly Disagree  $\bigcirc$  & $\bigcirc$ &  $\bigcirc$  &  $\bigcirc$  &  $\bigcirc$  Strongly Agree \\
    \hline
    \end{tabular}

    \item I thought the system was easy to use.
    \textit{Select only one bullet point.}\\
    \begin{tabular}{rcccl}
    1 & 2 & 3 & 4 & 5\\
    \hline
     Strongly Disagree  $\bigcirc$  & $\bigcirc$ &  $\bigcirc$  &  $\bigcirc$  &  $\bigcirc$  Strongly Agree \\
    \hline
    \end{tabular}

    \item  I think that I would need the support of a technical person to be able to use this system.
    \textit{Select only one bullet point.}\\
    \begin{tabular}{rcccl}
    1 & 2 & 3 & 4 & 5\\
    \hline
     Strongly Disagree  $\bigcirc$  & $\bigcirc$ &  $\bigcirc$  &  $\bigcirc$  &  $\bigcirc$  Strongly Agree \\
    \hline
    \end{tabular}

    \item  I found the various functions in this system were well integrated.
    \textit{Select only one bullet point.}\\
    \begin{tabular}{rcccl}
    1 & 2 & 3 & 4 & 5\\
    \hline
     Strongly Disagree  $\bigcirc$  & $\bigcirc$ &  $\bigcirc$  &  $\bigcirc$  &  $\bigcirc$  Strongly Agree \\
    \hline
    \end{tabular}

    \item I thought there was too much inconsistency in this system.
    \textit{Select only one bullet point.}\\
    \begin{tabular}{rcccl}
    1 & 2 & 3 & 4 & 5\\
    \hline
     Strongly Disagree  $\bigcirc$  & $\bigcirc$ &  $\bigcirc$  &  $\bigcirc$  &  $\bigcirc$  Strongly Agree \\
    \hline
    \end{tabular}

    \item I would imagine that most people would learn to use this system very quickly. 
    \textit{Select only one bullet point.}\\
    \begin{tabular}{rcccl}
    1 & 2 & 3 & 4 & 5\\
    \hline
     Strongly Disagree  $\bigcirc$  & $\bigcirc$ &  $\bigcirc$  &  $\bigcirc$  &  $\bigcirc$  Strongly Agree \\
    \hline
    \end{tabular}

    \item  I found the system very cumbersome to use.
    \textit{Select only one bullet point.}\\
    \begin{tabular}{rcccl}
    1 & 2 & 3 & 4 & 5\\
    \hline
     Strongly Disagree  $\bigcirc$  & $\bigcirc$ &  $\bigcirc$  &  $\bigcirc$  &  $\bigcirc$  Strongly Agree \\
    \hline
    \end{tabular}

    \item I felt very confident using the system.
    \textit{Select only one bullet point.}\\
    \begin{tabular}{rcccl}
    
    1 & 2 & 3 & 4 & 5\\
    \hline
     Strongly Disagree  $\bigcirc$  & $\bigcirc$ &  $\bigcirc$  &  $\bigcirc$  &  $\bigcirc$  Strongly Agree \\
    \hline
    \end{tabular}

    \item I needed to learn a lot of things before I could get going with this system.
    \textit{Select only one bullet point.}\\
    \begin{tabular}{rcccl}
    1 & 2 & 3 & 4 & 5\\
    \hline
     Strongly Disagree  $\bigcirc$  & $\bigcirc$ &  $\bigcirc$  &  $\bigcirc$  &  $\bigcirc$  Strongly Agree \\
    \hline
    \end{tabular}


    \item Mental Demand:How mentally demanding was the task?
    \textit{Please assign a score between 1 and 20, where 1 = very low, and 20 = very high.
}\\
    \underline{\qquad\qquad}

    \item Physical Demand:How physically demanding was the task?
    \textit{Please assign a score between 1 and 20, where 1 = very low, and 20 = very high.
}\\
    \underline{\qquad\qquad}

    \item Temporal Demand:How hurried or rushed was the pace of the task?
    \textit{Please assign a score between 1 and 20, where 1 = very low, and 20 = very high.
}\\
    \underline{\qquad\qquad}

    \item Performance: How successful were you in accomplishing what you were asked to do?
    \textit{Please assign a score between 1 and 20, where 1 = very low, and 20 = very high.
}\\
    \underline{\qquad\qquad}

    \item Effort:How hard did you have to work to accomplish your level of performance?
    \textit{Please assign a score between 1 and 20, where 1 = very low, and 20 = very high.
}\\
    \underline{\qquad\qquad}

    \item Frustration:How insecure, discouraged, irritated, stressed and annoyed wereyou?
    \textit{Please assign a score between 1 and 20, where 1 = very low, and 20 = very high.
}\\
    \underline{\qquad\qquad}

    \item Main source of workload?
    \textit{Select only one bullet point.}\\
    \begin{tabular}{rl}
     $\bigcirc$ Mental Demand   & $\bigcirc$ Physical Demand
    \end{tabular}

    \item Main source of workload?
    \textit{Select only one bullet point.}\\
    \begin{tabular}{rl}
     $\bigcirc$ Temporal Demand   & $\bigcirc$ Performance
    \end{tabular}

    \item Main source of workload?
    \textit{Select only one bullet point.}\\
    \begin{tabular}{rl}
     $\bigcirc$ Effort   & $\bigcirc$ Frustration
    \end{tabular}

    \item Main source of workload?
    \textit{Select only one bullet point.}\\
    \begin{tabular}{rl}
     $\bigcirc$ Mental Demand   & $\bigcirc$ Temporal Demand
    \end{tabular}

    \item Main source of workload?
    \textit{Select only one bullet point.}\\
    \begin{tabular}{rl}
     $\bigcirc$ Effort  & $\bigcirc$ Physical Demand
    \end{tabular}

    \item Main source of workload?
    \textit{Select only one bullet point.}\\
    \begin{tabular}{rl}
     $\bigcirc$ Performance   & $\bigcirc$ Frustration
    \end{tabular}

    \item Main source of workload?
    \textit{Select only one bullet point.}\\
    \begin{tabular}{rl}
     $\bigcirc$ Effort  & $\bigcirc$ Mental Demand
    \end{tabular}

    \item Main source of workload?
    \textit{Select only one bullet point.}\\
    \begin{tabular}{rl}
     $\bigcirc$ Temporal Demand   & $\bigcirc$ Frustration
    \end{tabular}

    \item Main source of workload?
    \textit{Select only one bullet point.}\\
    \begin{tabular}{rl}
     $\bigcirc$ Physical Demand  & $\bigcirc$ Performance
    \end{tabular}

    \item Main source of workload?
    \textit{Select only one bullet point.}\\
    \begin{tabular}{rl}
     $\bigcirc$ Mental Demand  & $\bigcirc$ Performance
    \end{tabular}

    \item Main source of workload?
    \textit{Select only one bullet point.}\\
    \begin{tabular}{rl}
     $\bigcirc$ Temporal Demand  & $\bigcirc$ Effort
    \end{tabular}

    \item Main source of workload?
    \textit{Select only one bullet point.}\\
    \begin{tabular}{rl}
     $\bigcirc$ Frustration   & $\bigcirc$ Physical Demand
    \end{tabular}

    \item Main source of workload?
    \textit{Select only one bullet point.}\\
    \begin{tabular}{rl}
     $\bigcirc$ Frustration & $\bigcirc$ Mental Demand
    \end{tabular}

    \item Main source of workload?
    \textit{Select only one bullet point.}\\
    \begin{tabular}{rl}
     $\bigcirc$ Frustration & $\bigcirc$ Temporal Demand
    \end{tabular}

    \item Main source of workload?
    \textit{Select only one bullet point.}\\
    \begin{tabular}{rl}
     $\bigcirc$ Performance & $\bigcirc$ Effort
    \end{tabular}

\end{enumerate}

\section{User study Details}
\label{User study Details}

\subsection{Participant Recruitment}
\label{Participant Recruitment}
We strategically recruited volunteers to participate in our user study, encompassing potential users of AutoML frameworks, consisting of 20 diverse participants: 6 software developers, 10 AI researchers, and 4 students, ensuring a rich blend of perspectives among the involved users.
Their prior exposure to key techniques relevant to this study is summarized below:
\begin{itemize}
    \item \textbf{Large Language Models:}
    \begin{itemize}
        \item 10 participants were very familiar with LLMs from actively using them in research projects.
        \item 5 participants had some previous experience with LLMs.
        \item 5 participants were completely new to LLMs. 
    \end{itemize}

    \item \textbf{AutoML \& HPO Tools:}
    \begin{itemize}
        \item 3 participants actively used AutoML \& HPO libraries like AutoGluon and Optuna in their work.
        \item 4 participants had tried basic AutoML tutorials before.
        \item 13 participants had no familiarity with AutoML and HPO tools.
    \end{itemize}

    \item \textbf{Multimodal Data Experience:}
    \begin{itemize}
        \item 5 participants worked extensively on multimodal datasets combining image, text, and tabular sources.
        \item 10 participants only used uni-modal datasets before.
        \item 5 participants were new to both multi-modal and uni-modal data.
    \end{itemize}
\end{itemize}

We believe that this diverse group of participants provides a comprehensive evaluation of our \texttt{AutoM\textsuperscript{3}L}, considering a range of backgrounds and expertise levels in AutoML methods. 

\subsection{Definition and Calculation of Variables}
\label{Definition and Calculation of Variables}

We denote participant responses to the \texttt{i\textsuperscript{th}} question in questionnaire \ref{Questionnaire After Task Execution} as \texttt{s\textsubscript{i}}. Questions 1-18 are numerical variables, while the remaining are categorical. The dependent variables are as follows:
\begin{itemize}
\item \textbf{Task Execution Time}: Objective, continuous variable measuring the total time taken by participants to successfully complete the task. Derived directly from the response to question 1 in questionnaire \ref{Questionnaire After Task Execution}:
\begin{equation}
\begin{aligned}
&\texttt{Time} = \texttt{s\textsubscript{1}}.
\end{aligned}
\end{equation}

\item \textbf{Number of Attempts}: Objective, continuous variable recording the total script execution attempts by participants to successfully complete the task. Derived directly from the response to question 2 in questionnaire \ref{Questionnaire After Task Execution}:
\begin{equation}
\begin{aligned}
\texttt{Attempts} &=  \texttt{s\textsubscript{2}}.
\end{aligned}
\end{equation}

\item \textbf{Usability Score}: 
The Usability Score is a subjective, continuous metric gauging the system's perceived usability. 
It is sourced from the System Usability Scale (SUS) survey questionnaire\cite{brooke1996sus}, which comprises 10 questions.
Each question offers five response choices from ``Strongly Disagree'' to ``Strongly Agree'', which are numerically scored from 1 to 5. 
Formally, the usability score is based on the responses to questions 3 through 12 in questionnaire \ref{Questionnaire After Task Execution}. 
To quantify usability, we apply the standard scoring system of the SUS to convert the scores for each participant on these questions into a new numerical format. 
Subsequently, we calculate the sum of these scores and multiply the result by 2.5. 
This step serves to reposition the original scores, which originally ranged from 0 to 40, into a revised scale spanning from 0 to 100. 
Although interpreted like percentiles, they aren't percentages.  
Higher scores signify better-perceived usability, which is mathematically defined as:
\begin{equation}
\begin{aligned}
\texttt{Usability} &= 2.5 \times \sum_{i=1}^{5}{(\texttt{s\textsubscript{1+2i}}-1) + (5 -\texttt{s\textsubscript{2+2i}})}.
\end{aligned}
\end{equation}

\item \textbf{Workload Index}: This subjective, continuous variable assesses perceived mental workload and is derived from the NASA Task Load Index (NASA TLX) questionnaire~\cite{hart1988development}. 
Recognized for its comprehensive evaluation of mental workload, the NASA TLX divides workload into six categories: Mental Demand, Physical Demand, Temporal Demand, Performance, Effort, and Frustration.
Participants rate each category on a scale of 1 to 20 (questions 13 to 18). They also evaluate the significance of 15 pairs of these categories in shaping the overall workload (questions 19 to 33).
The scale score for each dimension is calculated as $\texttt{s\textsubscript{i}} \times 5$. The weighted score \texttt{w\textsubscript{i}} is determined based on the frequency of selection for each dimension as more important in questions 19 to 33, divided by 15. 
The overall workload score, ranging from 0 to 100, is then computed by summing the products of the scale score and weighted score for each dimension as follows:
\begin{equation}
\begin{aligned}
\texttt{Workload} &= \sum_{i=13}^{18}{\texttt{w\textsubscript{i}}\cdot (5 \cdot \texttt{s\textsubscript{i})}}.
\end{aligned}
\end{equation}
\end{itemize}

The most direct independent variables stem from the differences in approaches between participants using \texttt{AutoM\textsuperscript{3}L} and AutoGluon when performing tasks. Furthermore, various independent variables have the potential to impact user outcomes, including:
\begin{itemize}
    \item \textbf{Participant Background:} These categorical variables encompass background information about the participants, such as their professional roles, providing deeper insights into potential background knowledge, biases, or preferences that users may bring to task execution.
    \item \textbf{Familiarity with Technology:} These numerical variables represent each participant's familiarity with terminal operations, the Python programming language, LLM, and AutoML methods. Familiarity levels can potentially impact the ease with which participants complete AutoML tasks, thus influencing the final measurement outcomes.
\end{itemize}

\subsection{User Study Analysis Process} 
\label{Analysis Process}
\begin{itemize}
\item \textbf{Collected Data.}
We collected both objective and subjective evaluations from each user regarding the systems, including task execution time, number of attempts, usability scores, and workload indices. Box plots for these four variables are presented individually in Fig \ref{boxplots}. Each box plot displays the minimum value, first quartile (Q1), median (Q2), third quartile (Q3), and maximum value for these variables. The box represents the interquartile range (IQR) from Q1 to Q3, with a line inside indicating the median. 

\begin{figure*}[ht!]
    \centering
    \includegraphics[width=0.50\linewidth]{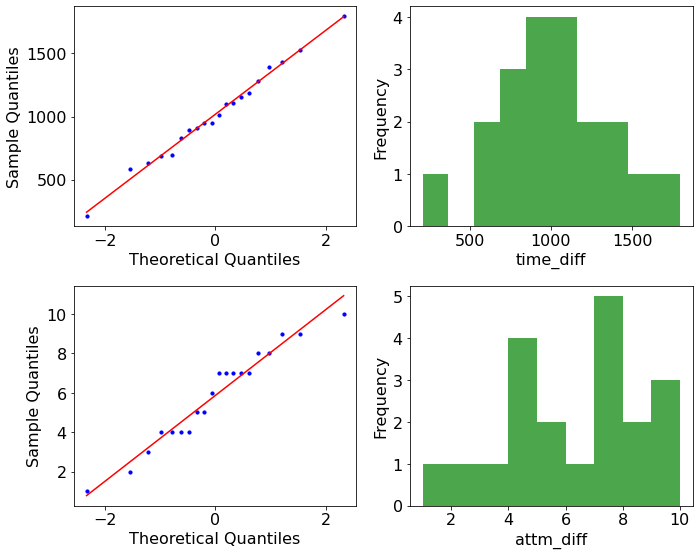}
    \caption{Normality Testing for task completion time and the number of attempts. The top row of panels present the Q-Q plot and histogram for task completion time, respectively. Similarly, the lower row of panels illustrate the Q-Q plot and histogram for the number of attempts.}
    \label{norm_test1}
\end{figure*}

\begin{figure*}[ht!]
    \centering
    \includegraphics[width=0.50\linewidth]{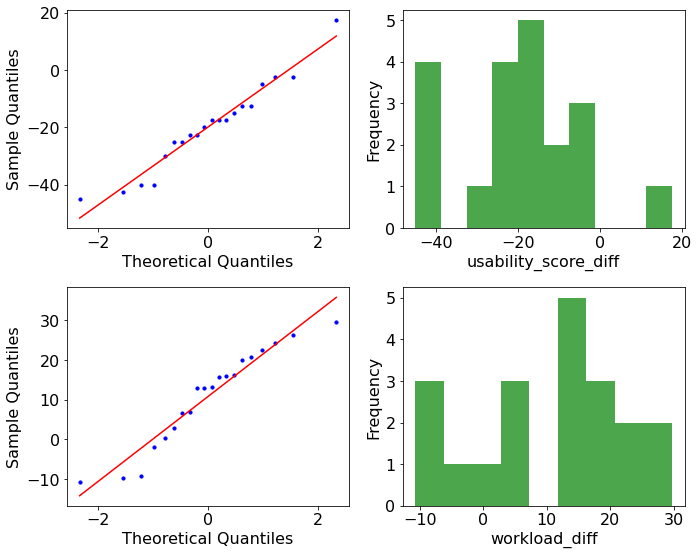}
    \caption{Normality Testing for system usability and workload. The top row of panels present the Q-Q plot and histogram for the usability, respectively. Similarly, the lower row of panels illustrate the Q-Q plot and histogram for the workload.}
    \label{norm_test2}
\end{figure*}

\item \textbf{Normality Testing.}
To ensure the validity of our subsequent statistical analyses, we conducted a normality test on our data using Q-Q plots, as depicted in Fig \ref{norm_test1} and Fig \ref{norm_test2}. The proximity of our data points to the theoretical quantile lines, along with the bell-shaped curve observed in the histograms, suggests that task completion time, the number of attempts, usability score and workload reasonably adhere to the assumption of normality.

\item \textbf{Hypothesis Testing.}
We employed hypothesis testing to assess the statistical significance of the observed performance differences between the AutoGluon and \texttt{AutoM\textsuperscript{3}L} conditions. The differences we are analyzing, denoted as \texttt{d\textsubscript{i}}, were calculated by taking the AutoGluon measurements and subtracting the corresponding \texttt{AutoM\textsuperscript{3}L}  measurements. Assuming the null hypothesis, both AutoGluon and \texttt{AutoM\textsuperscript{3}L} exert an equivalent impact. Consequently, these differences are expected to adhere to a distribution centered around zero, denoted as \texttt{µ\textsubscript{d}} = 0. Our dataset for hypothesis testing comprises 20 samples, and we express the null and alternative hypotheses as follows:

\begin{equation}
\begin{aligned}
\texttt{H\textsubscript{0}}:\texttt{µ\textsubscript{d}} &= 0  \quad \texttt{against} \quad \texttt{H\textsubscript{1}}:\texttt{µ\textsubscript{d}} \textgreater 0
\end{aligned}
\end{equation}

This applies to the testing of hypotheses H1, H2, and H4. In the case of testing H3, the alternative hypothesis is that $\texttt{µ\textsubscript{d}} \textless 0.$
Here, $ \overline{d}$ and $\texttt{s\textsubscript{d}}$ denote the sample mean and sample standard deviation of the observed differences, respectively. With these parameters in mind, the sampling distribution of the test statistic follows a t-distribution with degrees of freedom equal to n-1. Consequently, under the null hypothesis \texttt{H\textsubscript{0}},

\begin{equation}
\begin{aligned}
\tau &= \frac{\overline{d}}{\texttt{s\textsubscript{d}}/\sqrt{n}} \sim \texttt{t\textsubscript{n-1}}
\end{aligned}
\end{equation}

\end{itemize}

\section{Experiment Implementation}
\label{Experiment Implementation}
\subsection{Implementations for Quantitative Evaluations}
In our quantitative assessment, we primarily relied on  OpenAI’s APIs: \texttt{gpt-4-0314}\cite{openai2023gpt4} for code generation, and the \texttt{gpt-3.5-turbo\\-0301}\cite{gpt3.5} for text completion, and \texttt{text-embedding-ada-002}\cite{text-embedding} for text embedding. 
For all APIs, we set the temperature parameter to 0 to maximize determinism.
The experiments utilized the PyTorch Lightning framework\cite{falcon2019pytorch} for model training, and Ray\cite{moritz2018ray} served as our tool for hyperparameter search. 
Furthermore, we set the mode selection of AutoGluon to "best quality", ensuring optimal accuracy.
While we consistently used the same models in our multimodal experiments as those in the AutoGluon assessments for the same modality data, our emphasis on the model selection module was not solely on accuracy. Instead, we were driven by the goal of intelligently choosing models based on data modality and user-specific needs.
%
%
In order to obtain more robust experimental results, in all quantitative experiments, we used 10-fold cross-validation and reported the mean and standard deviation. For the retrieval dataset, stratified sampling of IDs is performed for each fold, and matching positive and negative sample pairs are created for verification.

\subsection{Implementations in User Study}
For the user study, given the advanced capabilities of GPT-3.5, we chose to employ the \texttt{gpt-3.5-turbo-0301} API as the LLM backbone of \texttt{AutoM\textsuperscript{3}L}. Participants in the study were provided execution scripts for both AutoGluon and \texttt{AutoM\textsuperscript{3}L}, allowing them a comparative experience.

\begin{table*}[t]
\centering
\caption{Robustness Assessment results of MS-LLM.}
\label{Robustness}
\begin{center}
\resizebox{\linewidth}{!}{
\begin{tabular}{|c|c|}
  \hline
  \textbf{id} & \textbf{sentences} \\
  \hline
  1 & I hope to see the model efficiently running on mobile devices, optimizing for lightweight performance. \\
  \hline
  2 & The model's deployment on CPU devices, especially on lightweight and mobile platforms, is my preference. \\
  \hline
  3 & My goal is to have the model effectively deployed on CPU devices, with a focus on mobile and lightweight configurations. \\
  \hline
  4 & It would be great to have the model running seamlessly on various CPU devices, prioritizing mobility and lightweight hardware. \\
  \hline
  5 & I'm aiming for the model to be deployed on specific CPU hardware, emphasizing mobility and lightweight characteristics. \\
  \hline
  6 & Optimizing the model for mobile platforms and ensuring efficient operation on CPU devices aligns with my preferences. \\
  \hline
  7 & The deployment of the model on CPU devices, particularly on lightweight and mobile configurations, is my desired outcome. \\
  \hline
  8 & I'm specifically interested in the model's deployment on CPU devices, emphasizing efficiency and suitability for mobile platforms. \\
  \hline
  9 & My preference is for the model to be tailored for deployment on CPU devices, with a keen focus on mobile and lightweight capabilities. \\
  \hline
  10 & Ensuring the model's inference speed on CPU devices, especially in mobile and lightweight scenarios, is my priority. \\
  \hline
  \multicolumn{2}{|c|}{\textbf{Results}} \\
  \hline
  \multicolumn{2}{|c|}{\{"google/flan-t5-small"; "mobilenetv3\_large\_100";  "categorical\_mlp"; "numerical\_mlp"\}} \\
  \hline
\end{tabular}
}
\end{center}
\end{table*}

\section{RELATED WORKS}
\subsection{AutoML}

AutoML has emerged as a transformative paradigm to streamline the design, training, and optimization of ML models by minimizing the need for extensive human intervention. 
Current AutoML solutions predominantly fall into three categories: (i) training pipeline automation, (ii) automated feature engineering, (iii) hyperparameter optimization.
Within the sphere of automated feature engineering, certain methodologies have carved a niche for themselves. 
For instance, DSM\cite{kanter2015deep} and OneBM\cite{lam2017one} have revolutionized feature discovery by seamlessly integrating with databases, curating an exhaustive set of features.
In a complementary vein, AutoLearn\cite{kaul2017autolearn} adopts a regression-centric strategy, enhancing individual records by predicting and appending additional feature values.
Concurrently, training pipeline and hyperparameter optimization automation have also seen significant advancements.
For example, H2O AutoML\cite{ledell2020h2o} is particularly noteworthy for its proficiency in rapidly navigating an expansive pipeline search space, leveraging its dual-stacked ensemble models.
%
%
However, a recurring challenge across these AutoML solutions is their predominant focus on uni-modal data, which limits their applicability to more complex multimodal data. 
At the same time, most of the current multimodal methods\cite{wang2019moc,wang2023deconfounded} designed for different application scenarios are not end-to-end and only support limited modality inputs, which limits the usability of the framework.
Recognizing this gap, we introduce a novel LLM framework tailored specifically for multimodal AutoML scenarios.

\subsection{Large Language Models}

The domain of Natural Language Processing has undergone a paradigm shift with the introduction of LLMs\cite{brown2020language,chowdhery2022palm,touvron2023llama,wei2022emergent,chung2022scaling}. 
With their staggering parameter counts reaching into the hundreds of billions, LLMs have showcased unparalleled versatility across diverse tasks. 
A testament to their evolving capabilities is Toolformer\cite{schick2023toolformer}, which equips LLMs to interact with external utilities via API calls, thereby expanding their functional horizons.
AutoGPT further exemplifies this evolution, segmenting broad objectives into tangible sub-goals, subsequently executed through prevalent tool APIs, such as search engines or code executors.
Yet, as we embrace the potential of LLMs to manage AI tasks via API interactions, it's crucial to navigate the inherent intricacies. 
Model APIs, in particular, often require bespoke implementations, frequently involving pre-training phases which highlights the pivotal role of AutoML in refining and optimizing these intricate workflows.
Our proposed AutoML framework aspires to bridge this gap, enabling fluid user-AI engagements through lucid dialogues and proficient code generation.

\section{Discusion}
\subsection{Biases and fragileness in LLMs}
Large lauguage models may contain biases that influence system performance and fairness. For instance, LLMs might exhibit gender or racial biases, leading to discriminatory outcomes during training and testing phases. We speculate the bias issues in LLMs may impact the \texttt{AutoM\textsuperscript{3}L}'s Automatic Feature Engineering module. For example, we expect LLM to identify and select skill names (attributes) relevant to job requirements. If LLM encounters bias in the training data, it might result in the following issues:
\begin{itemize}
    \item \textbf{Gender Bias:} The model might be inclined to select skill names associated with a specific gender, overlooking other equally important skills. For instance, there could be a tendency to select skills related to roles like engineers or programmers, neglecting skills required for roles such as nurses or educators.
    \item \textbf{Industry Bias:} The model might favor selecting skill names commonly used in that industry, neglecting skills required in other industries. This could lead to an imbalance in attribute selection across diverse industries.
\end{itemize}

To mitigate this issue, we propose:
\begin{itemize}
    \item \textbf{Fine-tuning LLMs:} Users should be aware of potential biases in LLMs and take measures to mitigate them. The selection of training data is crucial, ensuring that the dataset is diverse, inclusive, and covers various aspects such as gender, race, and culture.
    \item \textbf{Review and Correct Output of \texttt{AutoM\textsuperscript{3}L} Module:} Setting rules or adding post-processing steps to ensure the generated results do not contain adverse biases.
    \item \textbf{Improving Prompt Engineering for \texttt{AutoM\textsuperscript{3}L} Module:} For example, in the prompts, include diverse examples covering different genders, races, and industries. More specifically, design prompts relevant to the task's specific context to guide the model in better understanding and selecting attribute names. Additionally, reduce bias impact by introducing positive and negative examples. For instance:
    \begin{itemize}
        \item Positive Examples: Include positive examples related to various professions and skills, such as "programming," "project management," "communication skills," etc.
        \item Negative Examples: Introduce some irrelevant or inappropriate attributes, such as "gender," "appearance," "age," etc. These attributes are ones that I hope the model can ignore.
    \end{itemize}
    
\end{itemize}

\subsection{Computation cost}
The entire process of multimodal AutoML consists of two parts: setting up the training pipeline and the actual training of the model.
Our proposed LLM framework only participates in the setup phase. Once training begins, LLM is no longer used.
The majority of computational resources are consumed by the training itself, which remains unchanged regardless of LLM usage. 
Although LLM can introduce additional cost during the setup phase, this cost is minimal and has a negligible impact on the overall computational expense.
%
Moreover, the current API calls of LLM allow for real-time performance, making the training duration comparable to traditional methods.
For example, using dataset PAP, we conducted experiments on an RTX 3090 with identical settings for AutoM\textsuperscript{3}L and AutoGluon (same model and number of hyperparameter search trials).
Our training took 18.4 hours, while AutoGluon's took 18.3 hours, indicating that LLM only added 0.5\% to the total training time.
%
From an economic perspective, this experiment consumed approximately 200K tokens, costing no more than \$0.3 based on the current GPT-3.5 API pricing. 
%
%
Thus, incorporating LLMs does not result in a significant economic burden.

\subsection{Different LLMs}
We compared the use of GPT-4 and GPT-3.5-turbo in our framework, except for the code generation module. 
In a single trial on the PAP dataset, AutoM\textsuperscript{3}L w/GPT4 achieved an accuracy of 0.441, compared to AutoM\textsuperscript{3}L w/GPT3.5’s 0.439.
AutoM\textsuperscript{3}L w/ GPT4 demonstrated higher accuracy, which we attribute to GPT-4's superior comprehension and reasoning abilities.

\subsection{Robustness of model selection}
We employed the model zoo of HuggingFace Transformer, TIMM including different sizes of models for each modality.
As shown in Table \ref{Robustness}, we generated 10 different commands from a user instruction using GPT-3.5, all of which were accurately mapped to the lightweight models, indicating the robustness of MS-LLM.

\subsection{Interpretability}
AutoM\textsuperscript{3}L offers an interactive interface that allows users to actively participate in the pipeline construction process and receive explicit feedback at each stage.
During the model selection and hyperparameter optimization, users receive feedback from the LLM, 
such as why the selected model card matches user requirements, and the rationale behind the chosen hyperparameters, which provides a certain degree of transparency. 

\end{document}